\documentclass[10pt,letterpaper]{article}
\usepackage[T1]{fontenc}
\usepackage[utf8]{inputenc}
\usepackage{lmodern}
\usepackage[letterpaper,left=1in,right=1in,top=1.15in,bottom=.85in,
  headheight=35pt,headsep=16pt,footskip=28pt]{geometry}
\usepackage{microtype}
\usepackage{amsmath,amssymb}
\usepackage{graphicx,xcolor}
\usepackage{booktabs,array,tabularx,makecell}
\usepackage{algorithm,algpseudocode}
\usepackage{float,placeins,needspace}
\usepackage[font=small,labelfont=bf]{caption}
\usepackage[authoryear,round]{natbib}
\setcitestyle{authoryear,round,citesep={;},aysep={,},yysep={;}}
\usepackage{url}
\usepackage{fancyhdr}
\usepackage[hidelinks,hyperfootnotes=false]{hyperref}
\hypersetup{
  pdfauthor={Haoran Lang; Haotao Lu; Shiyu Sang; Haoyang Luo; Guo Chen; Qun Li; Jingyi Yu; Ye Shi; Jingya Wang},
  pdftitle={EmbodiRSI: Recursive Self-Improvement for Data-Efficient Robot Adaptation},
  pdfsubject={Robot adaptation and recursive self-improvement},
  pdfkeywords={robot learning, recursive self-improvement, real-to-sim-to-real, data-efficient adaptation}
}

\newcolumntype{L}[1]{>{\raggedright\arraybackslash}p{#1}}
\newcolumntype{C}[1]{>{\centering\arraybackslash}p{#1}}

\newcommand{\NA}{\textemdash}

\fancypagestyle{titlepage}{%
  \fancyhf{}%
  \fancyhead[L]{\includegraphics[width=1.18in]{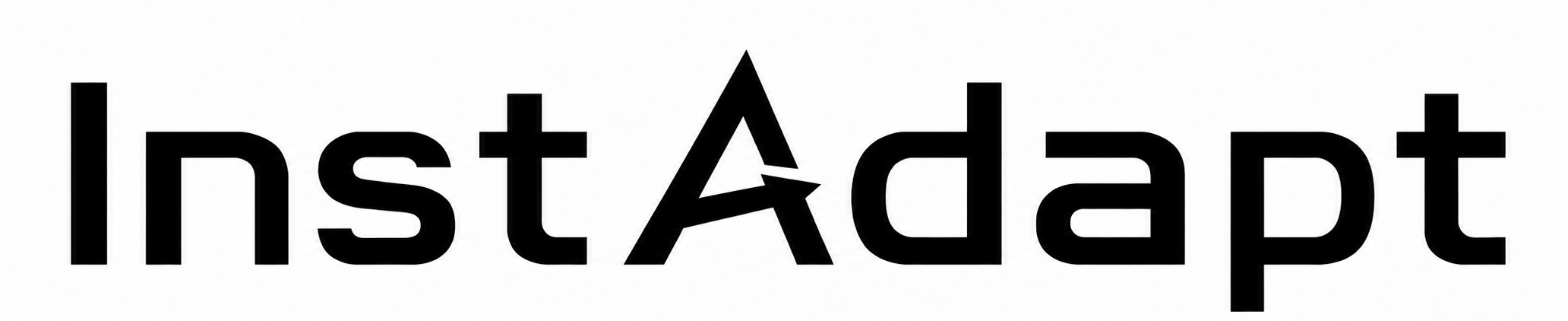}}%
}

\newcommand{\AuthorEntry}[2]{\mbox{#1\textsuperscript{\normalfont\ensuremath{#2}}}}
\newcommand{\AuthorEmail}[1]{\href{mailto:#1@shanghaitech.edu.cn}{#1}}

\makeatletter
\renewcommand{\maketitle}{%
  \thispagestyle{titlepage}%
  \begingroup
  \centering
  \setlength{\parskip}{0pt}%
  \vspace*{7pt}%
  {\fontsize{17}{21}\selectfont\bfseries\@title\par}%
  \vspace{17pt}%
  {\fontsize{11}{15}\selectfont\bfseries\@author\par}%
  \vspace{12pt}%
  {\fontsize{10}{13}\selectfont\PaperAffiliations\par}%
  \vspace{10pt}%
  {\fontsize{10}{13}\selectfont
    \href{https://embodirsi.instadapt.com}{\textcolor{blue}{EmbodiRSI Project Page}}\par}%
  \vspace{8pt}%
  {\fontsize{8.6}{11}\selectfont\ttfamily\PaperEmails\par}%
  \vspace{9pt}%
  \endgroup
}
\renewenvironment{abstract}{%
  \par\vspace{2pt}%
  \begin{center}\large\bfseries Abstract\end{center}%
  \begin{list}{}{\setlength{\leftmargin}{.45in}\setlength{\rightmargin}{.45in}}%
  \item\relax\small
}{%
  \end{list}\vspace{6pt}%
}
\renewcommand{\section}{\@startsection{section}{1}{\z@}%
  {-2.4ex plus -.5ex minus -.2ex}{1.2ex plus .2ex}%
  {\large\bfseries\raggedright}}
\renewcommand{\subsection}{\@startsection{subsection}{2}{\z@}%
  {-2ex plus -.4ex minus -.2ex}{.8ex plus .2ex}%
  {\normalsize\bfseries\raggedright}}
\renewcommand{\subsubsection}{\@startsection{subsubsection}{3}{\z@}%
  {-1.8ex plus -.4ex minus -.2ex}{.6ex plus .2ex}%
  {\normalsize\bfseries\raggedright}}
\renewcommand{\paragraph}{\@startsection{paragraph}{4}{\z@}%
  {1.5ex plus .3ex minus .2ex}{-1em}{\normalsize\bfseries}}
\makeatother

\title{EmbodiRSI: Recursive Self-Improvement\\
for Data-Efficient Robot Adaptation}

\author{%
  \AuthorEntry{Haoran Lang}{1,2,*}\quad
  \AuthorEntry{Haotao Lu}{1,2,*}\quad
  \AuthorEntry{Shiyu Sang}{1,2,*}\quad
  \AuthorEntry{Haoyang Luo}{1}\\[5pt]
  \AuthorEntry{Guo Chen}{1}\quad
  \AuthorEntry{Qun Li}{1}\quad
  \AuthorEntry{Jingyi Yu}{1}\quad
  \AuthorEntry{Ye Shi}{1,2,\dagger}\quad
  \AuthorEntry{Jingya Wang}{1,2,\dagger}%
}
\newcommand{\PaperAffiliations}{%
  \textsuperscript{1}ShanghaiTech University\qquad
  \textsuperscript{2}InstAdapt%
}
\newcommand{\PaperEmails}{%
  \{\AuthorEmail{langhr2025},\AuthorEmail{luht2025},\AuthorEmail{sangshy2025},%
  \AuthorEmail{luohy12024},\AuthorEmail{chenguo2024},\\[2pt]
  \AuthorEmail{liqun2024},\AuthorEmail{yujingyi},\AuthorEmail{shiye},%
  \AuthorEmail{wangjingya}\}@shanghaitech.edu.cn%
}
\date{}

\begin{document}
\maketitle
\begingroup
  \renewcommand{\thefootnote}{\fnsymbol{footnote}}
  \footnotetext[1]{These authors contributed equally to this work.}
  \footnotetext[2]{Corresponding authors: Jingya Wang
    (\href{mailto:wangjingya@shanghaitech.edu.cn}{\texttt{wangjingya@shanghaitech.edu.cn}}), 
    Ye Shi
    (\href{mailto:shiye@shanghaitech.edu.cn}
    {\texttt{shiye@shanghaitech.edu.cn}}).}
\endgroup
\setcounter{footnote}{0}
\begin{figure}[H]
\centering
\includegraphics[width=\linewidth]{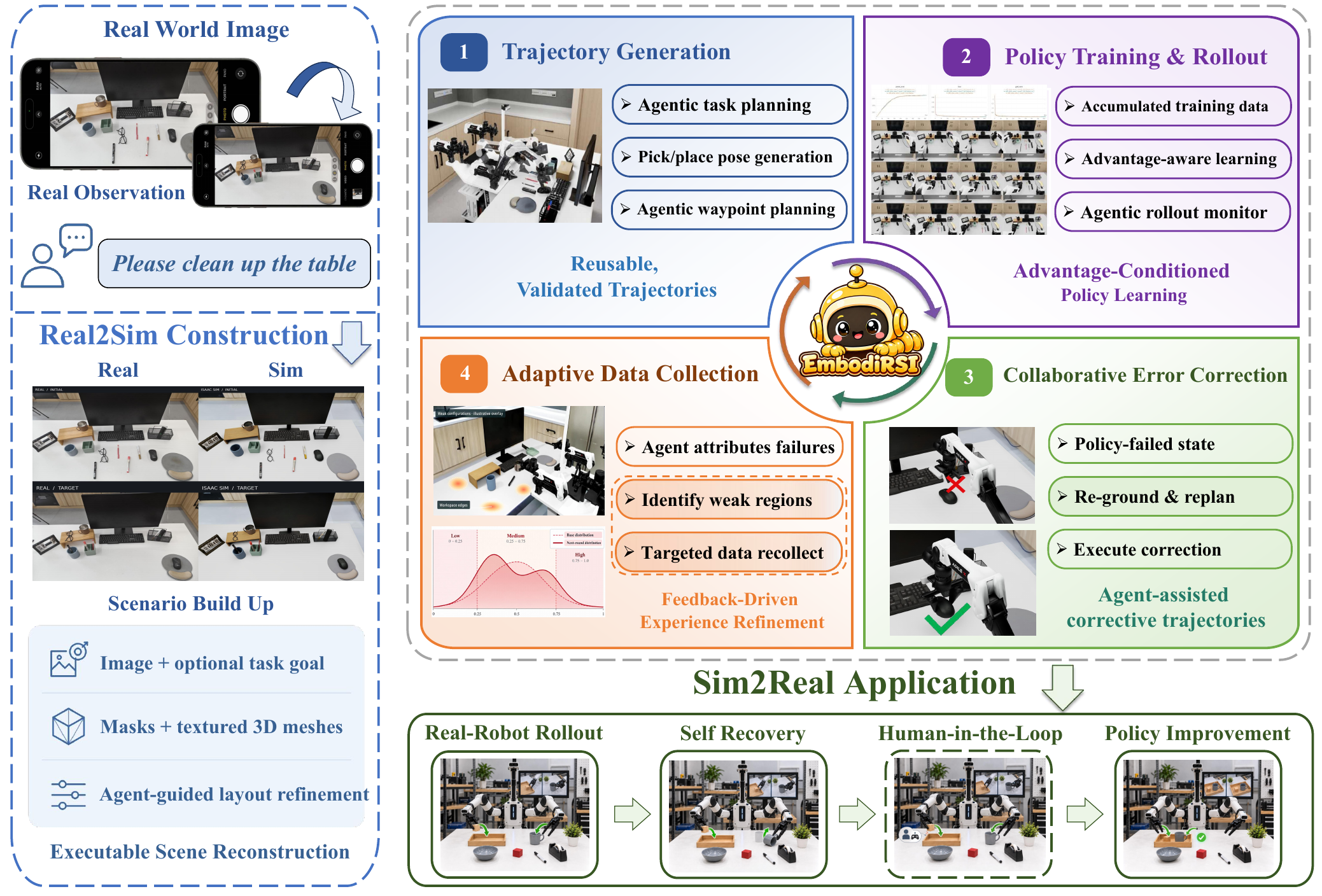}
\caption{\textbf{Overview of EmbodiRSI.}
Real-world observations and task goals guide the construction of an executable
simulation environment and the generation of validated demonstrations.
Within this environment, policy rollout feedback guides Collaborative Error
Correction and Adaptive Data Collection.
The resulting experience updates the policy for the next round,
forming a recursive self-improvement loop.
A small real-world human-in-the-loop (HIL) dataset then refines the policy
for autonomous deployment.}

\label{fig:embodirsi_overview}
\label{fig:overview}
\end{figure}

\clearpage

\begin{abstract}
Adapting robot manipulation policies to new tasks and environments remains highly data-intensive, while the data needed for further improvement depends on the policy's current capabilities and failure modes. We introduce EmbodiRSI, an agentic system for recursive self-improvement (RSI) in a real-to-sim-to-real setting, where task-specific simulations are constructed from target deployment scenarios and used as low-cost environments for iterative policy improvement before transfer back to the physical world. EmbodiRSI uses policy execution feedback to guide subsequent experience acquisition and policy updates. Two complementary mechanisms close this loop: \emph{Collaborative Error Correction} generates agent-assisted corrective trajectories from policy-reached states, while \emph{Adaptive Data Collection} directs expert demonstration generation toward the current policy's weaknesses. The task-specific simulation serves as a reusable workspace for policy warm-up, repeatable evaluation, failure diagnosis, and targeted data generation across successive RSI rounds. Across three tabletop environments and 14 subtasks, EmbodiRSI increases scene-balanced autonomous simulation success from 50.4\% to 83.5\% over two RSI updates. With 400 adaptive simulated trajectories and only ten real-world refinement trajectories per subtask, EmbodiRSI achieves 83.1\% scene-balanced autonomous real-world success, compared with 75.0\% for adaptation using 200 real-world demonstrations per subtask. These results demonstrate that feedback-driven recursive improvement in deployment-specific simulations can enable data-efficient adaptation of embodied policies to physical environments.
\end{abstract}

\section{Introduction}
\label{sec:introduction}
Recent advances in general-purpose robot manipulation policies have substantially expanded the range of tasks that a single robot can perform~\citep{kim2024openvla,black2024pi_0,intelligence2025pi_}, yet reliable deployment in new environments remains challenging. Deployment often reveals systematic failure modes that are insufficiently covered by the original training distribution. Importantly, these failure modes change as the policy improves: some weaknesses are resolved, while the remaining errors become concentrated in different states or configurations. Effective adaptation therefore requires more than a fixed training set. It calls for a learning system that repeatedly evaluates the current policy, identifies its remaining weaknesses, and acquires targeted experience for subsequent improvement, without relying on costly physical data collection and human intervention at every iteration.

The missing link is between \emph{observing a failure} and \emph{obtaining useful
experience for subsequent improvement}. Within an episode, a missed grasp or displaced object can leave
the robot in a state absent from nominal demonstrations. Across episodes, recurring failures reveal regions of the task distribution where additional experience is most needed. Dataset aggregation supervises learner-induced
states~\citep{ross2011reduction}; intervention and recovery methods turn execution
errors into training experience~\citep{hoque2024intervengen,lin2025failsafe}; and recent agentic systems coordinate collection, learning and execution~\citep{li2026roboclaw}. These directions motivate a broader question:

\emph{Can a robot use feedback from its current policy to determine both how to recover from failures and what experience should be acquired for the next policy update?}

We introduce \textbf{EmbodiRSI}, an agentic learning system for \emph{recursive self-improvement} (RSI) within a \emph{real-to-sim-to-real} (R2S2R) workflow
(Figure~\ref{fig:overview}). Starting from a target physical workspace, EmbodiRSI constructs a task-specific simulation in which the policy can be repeatedly evaluated and improved before transferring back to the real world. The current policy generates rollouts, execution outcomes determine what experience should be acquired next, and the resulting experience is used to update the policy. The updated policy is then evaluated again, producing new feedback for the next round. In this way, the experience distribution evolves together with the policy rather than being fixed in advance.

Two complementary mechanisms close this RSI loop. \emph{Collaborative Error Correction} operates within an episode: when policy execution fails, an agent guides recovery from the intermediate state actually reached by the policy, generating corrective trajectories without resetting the task. \emph{Adaptive Data Collection} operates across episodes: the agent aggregates rollout outcomes, identifies recurring failure modes, and redirects subsequent expert demonstration generation toward configurations where the current policy remains weak. Together, these mechanisms transform execution feedback into targeted training experience for subsequent policy updates. The updated policy is then rolled out again to reveal its remaining limitations.

Task-specific simulation provides the workspace in which this recursive improvement can be performed efficiently and repeatedly. Given observations of the target deployment environment and task specifications, EmbodiRSI reconstructs and organizes a corresponding virtual workspace, while a trajectory agent generates validated expert demonstrations for policy initialization and subsequent adaptive collection. The same environment supports repeatable evaluation, failure diagnosis, error correction, and targeted data generation. Geometric reasoning, collision checking, and motion-planning tools are used to validate generated trajectories and recovery actions. After iterative improvement in simulation, only a small number of human-in-the-loop (HIL) trajectories are collected in the physical workspace to refine the policy for autonomous real-world execution.

Our evaluation follows this R2S2R progression from recursive improvement in task-specific simulation to deployment in the corresponding physical environments. Across three tabletop environments and 14 subtasks, EmbodiRSI increases scene-balanced autonomous simulation success from 50.4\% to 83.5\% over two RSI updates (R0 to R2). To isolate the benefit of feedback-driven experience acquisition, we further compare adaptive and static collection in two environments at matched retained-trajectory counts. With a cumulative budget of 400 simulated trajectories per subtask, adaptive collection improves scene-balanced simulation success by 20.6 percentage points and pre-refinement real-world success by 12.9 points over static collection. In the full three-environment evaluation, EmbodiRSI achieves 83.1\% scene-balanced autonomous real-world success using 400 adaptive simulated trajectories and only ten real-world refinement trajectories per subtask, compared with 75.0\% for adaptation using 200 real-world demonstrations per subtask.

Our main contributions are as follows.
\begin{itemize}

\item We propose EmbodiRSI, an agentic recursive self-improvement system within an R2S2R workflow, where policy execution feedback iteratively guides experience acquisition and policy updates in task-specific simulation environments.

\item We develop two complementary mechanisms to close the RSI loop: \emph{Adaptive Data Collection}, which directs expert demonstrations toward the current policy's weaknesses, and \emph{Collaborative Error Correction}, which generates corrective trajectories from policy-reached failure states through agent-guided recovery.

\item We demonstrate progressive policy improvement and data-efficient real-world adaptation across diverse manipulation tasks. EmbodiRSI achieves higher real-world success with only a small amount of physical refinement data than adaptation using substantially more real-world demonstrations.

\end{itemize}

\section{Related Work}
\label{sec:related_work}

\paragraph{Simulation-based Scalable Data Generation.}
RoboGen and GenSim generate environments or tasks~\citep{wang2023robogen,wang2024gensim};
MimicGen, SkillMimicGen, DexMimicGen and DemoGen expand source
demonstrations~\citep{mandlekar2023mimicgen,garrett2024skillmimicgen,jiang2025dexmimicgen,xue2025demogen}. RoboTwin uses generative digital twins~\citep{mu2025robotwin} and RoboTwin~2.0 scales
bimanual data generation with domain randomization~\citep{chen2025robotwin}.
SAGE and V-CAGE incorporate agentic generation and
verification~\citep{xia2026sage,liu2026v}. Deployment-specific Real2Sim2Real systems connect reconstructed workspaces to
policy learning and transfer~\citep{torne2024reconciling,fang2025rebot,han2025re,dan2025x,patel2025real}.
 EmbodiRSI uses these capabilities as infrastructure for repeated policy
improvement: the workspace remains available after initial training, while rollout feedback revises what experience is generated within it.

\paragraph{Robot Learning via Real2Sim2Real.}
Real2Sim2Real reduces real-world data requirements by reconstructing or generating simulation from real observations, learning policies or expanding robot experience in simulation, and transferring the resulting policies or behaviors back to real robots. Representative systems include RialTo~\citep{torne2024reconciling}, RL-GSBridge~\citep{wu2025rl}, RoboGSim~\citep{li2024robogsim}, Re\(^3\)Sim~\citep{han2025re}, RoboSimGS~\citep{zhao2026high}, ReBot~\citep{fang2025rebot}, and X-SIM~\citep{dan2025x}. Closely related real-to-sim and photorealistic digital-twin systems, such as IKER~\citep{patel2025real}, Scalable Real2Sim~\citep{pfaff2025scalable}, RoboPearls~\citep{tao2025robopearls}, GSWorld~\citep{jiang2025gsworld}, and Real2Sim-Eval~\citep{zhang2025real}, demonstrate the value of simulation as a bridge between real observations and robot execution.  EmbodiRSI further extends this paradigm with an agent-driven closed-loop that unifies scene construction, trajectory generation, policy training, rollout evaluation, failure recovery, and data refinement.

\paragraph{Recursive Self-Improvement for Robot Learning.}
Recent work on recursive robot self-improvement explores how policies can be iteratively refined using experience generated by their own evolving behavior. SOAR~\citep{zhou2024autonomous} uses foundation models to guide autonomous data collection and learning from unlabeled experience, while RECAP~\citep{intelligence2025pi} combines demonstrations, autonomous experience, and corrective feedback through advantage conditioning. VERITAS~\citep{zhang2026visual} uses visual verification to filter autonomous rollouts for subsequent policy improvement, and RoboClaw~\citep{li2026roboclaw} integrates autonomous data collection, learning, and execution into a unified iterative loop. RISE~\citep{yang2026rise} further improves policies through imagined rollouts and advantage estimation in a learned world model.  EmbodiRSI follows this direction by coupling within-episode recovery with across-episode recollection, so that experience generated by each policy iteration informs subsequent updates.
Appendix~\ref{si-sec:related_work} provides further context.

\section{Method: A System for Recursive Policy Self-Improvement}
\label{sec:method}
 EmbodiRSI couples policy optimization to the acquisition of its next training
experience. At round $k$, let $\pi_k$ be the current scene-specific policy,
$\mathcal D_k$ the accumulated simulation dataset and $p_k(c)$ the distribution
of demonstration-collection configurations. A configuration $c$ specifies a
source object, target anchor, instruction, grasp and waypoint program, and
admissible reset ranges. One policy covers all subtasks in a scene. Sections
\ref{sec:scene_generation} and~\ref{sec:trajectory_main} describe the supporting
infrastructure; Section~\ref{sec:feedback} develops the RSI loop.

\subsection{An Executable Environment for RSI}
\label{sec3.1}\label{sec:scene_generation}
Given a tabletop image and, optionally, a target image or language goal,  EmbodiRSI constructs an initial scene $\mathcal S_0$ and an organized target scene $\mathcal S_{\mathrm{org}}$ through a fully automated, agent-orchestrated pipeline.
The scene agent identifies objects and invokes SAM~3~\citep{carion2026sam} to obtain instance masks and SAM~3D~\citep{chen2026sam} to reconstruct the corresponding textured meshes.
It then applies Agent-Guided Adaptive Collision Rectification to resolve interpenetration while preserving support and containment relations.
To construct the organized target scene, the agent establishes coarse task relations and refines object poses using rendered views and geometric checks.

This scene is the \emph{practice environment} for RSI: it supports repeatable resets, exposes object states for training-time diagnosis and permits recovery from policy-failed states. It is retained throughout learning, rather than used only to produce a seed dataset. Scene construction, geometric operators and organization examples are detailed in Appendices~\ref{si-sec:infrastructure} and~\ref{si-sec:tasks}.

\subsection{A Reusable Source of Training Experience}
\label{sec3.2}\label{sec:trajectory_main}
A trajectory agent compiles the scene pair into executable task configurations and validated demonstrations. It proposes object-space routes, projects them through GraspGen~\citep{murali2025graspgen} candidates, and validates complete robot programs with cuRobo~\citep{sundaralingam2026curobov2}.
Admission requires swept-object collision checking, multi-view review, sequential motion feasibility and execution-time task checks. Infeasible candidates are repaired and revalidated or rejected.

Stored waypoints follow source or target anchors. After a reset, they are
re-grounded to measured object poses and replanned from the current robot state.
This reusable interface connects agent decisions to executable supervision for
initial collection, recovery and targeted recollection.
Appendix~\ref{si-sec:trajectory_details} details compilation and execution.

\subsection{Agent-Guided Recursive Policy Self-Improvement}
\label{sec3.3}\label{sec:feedback}
\begin{figure}[t]
\centering
\includegraphics[width=\linewidth]{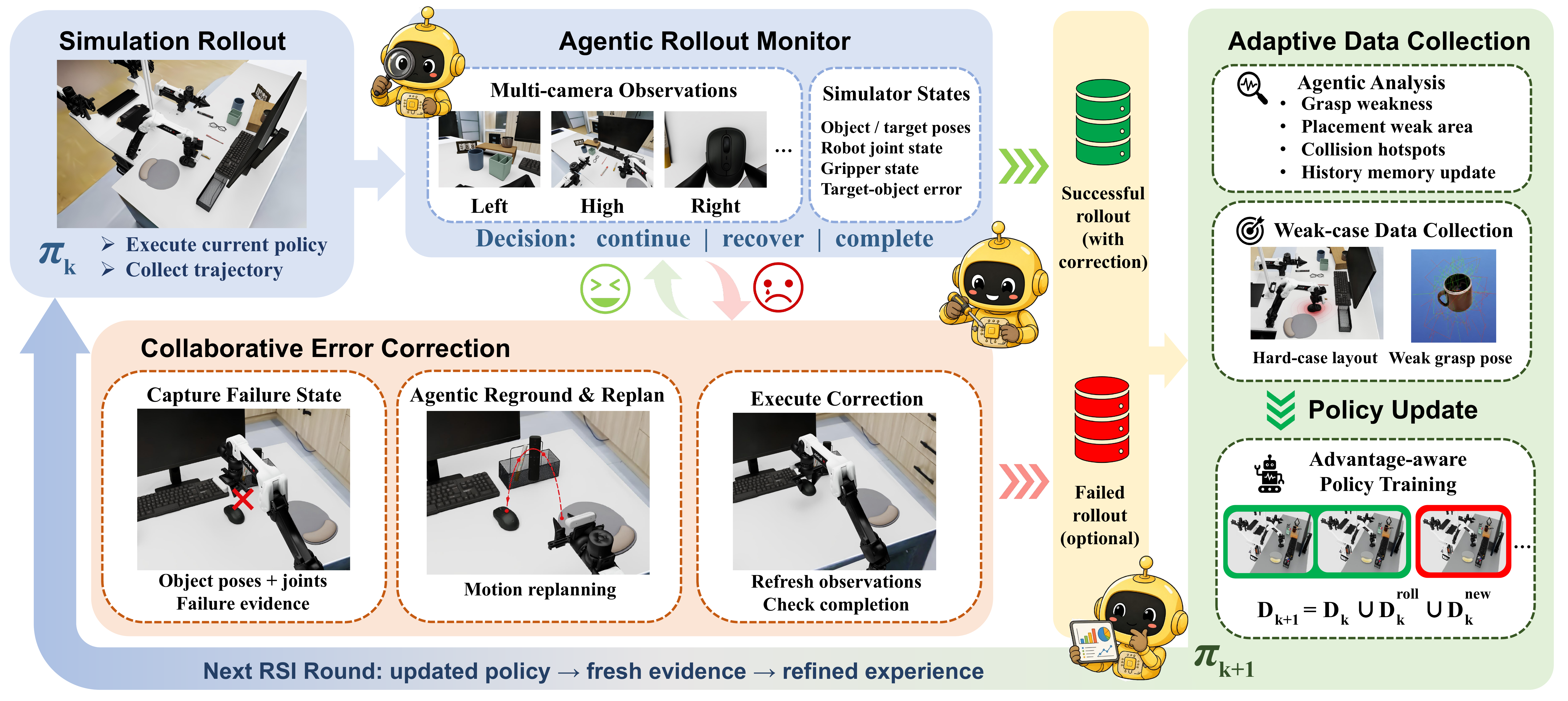}
\caption{\textbf{The recursive self-improvement loop in EmbodiRSI.}
The rollout monitor combines multi-camera observations and simulator states
to trigger recovery from the current state without resetting.
Failure analysis guides targeted data collection.
Successful rollouts, including those completed after correction, and newly
generated demonstrations are accumulated to update the policy for the next round.
External recovery is used only during training; evaluation is autonomous.}
\label{fig:recovery_pipeline}
\end{figure}

Initial demonstrations may omit both imperfect learner-induced states and
configurations that remain difficult after training.  EmbodiRSI addresses these
gaps at two timescales (Figure~\ref{fig:recovery_pipeline}): recovery supplies
corrective supervision \emph{within} an episode, while rollout diagnosis
refines the collection distribution \emph{across} episodes. Accumulated
experience updates the policy, closing the loop summarized in
Algorithm~\ref{alg:evolution}.

\subsubsection{Agentic Rollout Monitor}
\label{sec:monitoring}
During a rollout $\tau^{\pi_k}=\{(o_t,x_t,a_t)\}_{t=1}^{T}$, a monitor agent combines a temporal image window, action history and simulator object/robot state $x_t$:
\begin{equation}
 y_t=\mathcal A_{\mathrm{RSI}}(o_{t-w:t},x_t,a_{1:t},l),\qquad
 y_t\in\{\texttt{continue},\texttt{recover},\texttt{complete}\}.
 \label{eq:monitor}
\end{equation}
Visual evidence identifies missed grasps, slippage and incorrect placement;
programmatic checks measure target-object error and gripper state. Completion requires both a visual decision and geometric success. Recovery is gated by confidence, cooldown, attempt budget and an object-holding test, avoiding interruption of an ongoing valid transport.
Decisions are logged with configuration identifiers, scene states and failure evidence. Simulator object poses support these training tools but are not inputs to the learned policy.

\subsubsection{Collaborative Error Correction}
\label{sec:recovery}
Recovery must solve the task from the state the policy actually reached, not
replay a clean demonstration from its initial state. At a recovery trigger
$t_f$, a nominal grasp is transported to the current object pose:
\begin{equation}
 T_g^{\mathrm{cur}}=T_o^{\mathrm{cur}}(T_o^0)^{-1}T_g^0.
 \label{eq:recovery_grasp}
\end{equation}
Here, $T_o^{\mathrm{cur}}$ and $T_o^0$ denote the current and initial poses of object $o$, respectively.
Candidate configurations are probed in deterministic order without advancing the simulator.
Entry motion is planned from the \emph{current joint state}, and targets are re-grounded to the current scene.
The correction can provide an entry-to-grasp motion or continue the remaining manipulation program.

Diagnosis and planning pause simulation advancement. After a correction, the system refreshes observations, checks completion and may return control to the policy under the recovery gates.
We log recovery requests, executed corrections and final episode outcomes separately.
Successful continuations provide action supervision from imperfect states that nominal demonstrations may omit.
Recovery supplies corrective training experience from policy-failed states, while performance evaluation uses the learned policy alone.

\subsubsection{Adaptive Data Collection}
\label{sec:recollection}
Agent-guided data collection adapts the collection distribution to the current policy's diagnosed weaknesses. After each collection round, the agent attributes final failures to the source-object reset, target-anchor reset or neither; stored attributions can be reused without additional VLM calls.
For reset subject $s$ with a non-degenerate, configuration-specific range
$[\ell^s,h^s]$, we normalize its position to a common grid and compute
the attributed-failure rate:
\begin{equation}
 u^s(x,y)=\left(\frac{x-\ell_x^s}{h_x^s-\ell_x^s},
                 \frac{y-\ell_y^s}{h_y^s-\ell_y^s}\right),\qquad
 \widehat r_{k,b}^s=\frac{U_{k,b}^s}{N_{k,b}^s}.
 \label{eq:grid_attribution}
\end{equation}
Here, $N_{k,b}^s$ counts completed monitored episodes with valid subject positions in cell $b$, whereas $U_{k,b}^s$ counts only final failures with supported attribution to $s$. Cell selection requires sufficient observations and attributed failures, together with a rate exceeding the subject-level global rate. Episodes that succeed after recovery do not contribute to the numerator; these statistics describe assisted collection rather than unassisted test performance.

Selected cells are mapped back to world coordinates to restrict the
corresponding reset ranges, including yaw when sufficiently supported.
Configurations for subsequent demonstration generation are sampled from
\begin{equation}
 p_{k+1}(c)=(1-\alpha)p_0(c)+\alpha p_{\mathrm{hard},k}(c),
 \qquad 0\leq\alpha\leq1.
 \label{eq:collection_distribution}
\end{equation}

Here, $p_{\mathrm{hard},k}$ concentrates sampling on diagnosed weak
configurations, while the base component preserves access to the original range for $\alpha<1$.
The mixture controls configuration sampling but does not determine retained-trajectory proportions, which also depend on execution-based acceptance.
This refinement applies only to training collection; test resets remain unchanged.
Failure concentration provides a heuristic for prioritizing subsequent data acquisition (Appendix~\ref{si-sec:grid_attribution}).

\subsubsection{Policy Updating and Recursion}
\label{sec:policy_learning}
Let $\mathcal D_k^{\mathrm{roll}}$ contain retained rollout-derived episodes,
which may include both policy actions and corrections, and let
$\mathcal D_k^{\mathrm{new}}$ contain validated demonstrations generated under
$p_{k+1}$. Each saved episode is counted once. The next dataset is
\begin{equation}
 \mathcal D_{k+1}=\mathcal D_k\cup\mathcal D_k^{\mathrm{roll}}
                               \cup\mathcal D_k^{\mathrm{new}}.
 \label{eq:data_update}
\end{equation}
The policy receives head- and wrist-camera RGB images, a 14-dimensional robot
state (six arm-joint positions and one gripper value per arm) and an instruction.
Following RECAP~\citep{intelligence2025pi}, a task-conditioned value model annotates the accumulated experience:
\begin{equation}
 \widehat A_t^{(n)}=R_t^{(n)}+V_\phi(o_{t+n},l)-V_\phi(o_t,l),\qquad
 z_t=\mathbb I[\widehat A_t^{(n)}>\delta].
 \label{eq:td_advantage}
\end{equation}
The implementation sets $R_t^{(n)}=0$, $n=50$, so the binary condition depends on the value-estimate change.
With $\widetilde{\mathcal D}_{k+1}$ denoting the annotated data, the policy update minimizes
\begin{equation}
 \mathcal L_\pi(\theta)=
 \mathbb E_{(o_t,l,a_t,z_t)\sim\widetilde{\mathcal D}_{k+1}}
 \left[\ell_{\mathrm{BC}}\bigl(\pi_\theta(\cdot\mid o_t,l,z_t),a_t\bigr)\right].
 \label{eq:policy_learning}
\end{equation}
Recorded actions remain the supervision targets, while advantage labels are provided in text form as condition for action prediction.
This formulation integrates RSI-generated experience into policy learning. The updated policy $\pi_{k+1}$ then generates the next round's rollouts, which guide the refinement of both corrective supervision and subsequent data collection.
The number of rounds and retained-data quotas are fixed in advance; feedback guides experience acquisition within this schedule.

\begin{algorithm}[t]
\caption{ EmbodiRSI: recursive policy self-improvement}
\label{alg:evolution}
\begin{algorithmic}[1]
\Require Deployment observations and goals, base distribution $p_0$, round/data budgets
\State Construct the simulator; compile validated task configurations
\State Collect initial demonstrations $\mathcal D_0$ and train $\pi_0$
\For{each scheduled RSI round $k$}
\State Roll out $\pi_k$; diagnose execution and request eligible state-conditioned recovery
\State Retain rollout-derived episodes $\mathcal D_k^{\mathrm{roll}}$, counting each episode once
\State Attribute final failures; form $p_{k+1}$ by Eq.~\eqref{eq:collection_distribution}
\State Generate and validate new demonstrations $\mathcal D_k^{\mathrm{new}}$ under $p_{k+1}$
\State Accumulate $\mathcal D_{k+1}$; annotate and train $\pi_{k+1}$ by Eqs.~\eqref{eq:data_update}--\eqref{eq:policy_learning}
\EndFor
\State Refine the selected policy with a small real-world human-in-the-loop~(HIL) dataset
\State Evaluate independently with the learned policy alone
\end{algorithmic}
\end{algorithm}

\paragraph{Coordination and real-world adaptation.}
\label{sec:agent-supervision}
At the outer level, an OpenClaw-based supervisor coordinates the complete RSI workflow through a Model Context Protocol (MCP) interface. (Appendix~\ref{si-sec:supervisor_runtime}).
After simulation RSI, ten retained real-world human-in-the-loop~(HIL) trajectories per subtask refine the selected policy with the same learning formulation.
Operators correct residual execution errors, such as unstable grasps and inaccurate placement.
Final simulation and physical tests are independent of training collection and disable agent takeover, planner recovery and human correction.
A test-time reattempt is therefore behavior learned by the policy itself.

\section{Experiments}
\label{sec:results}\label{sec:experiments}

\begin{figure}[!t]
\centering
\includegraphics[width=.94\linewidth]{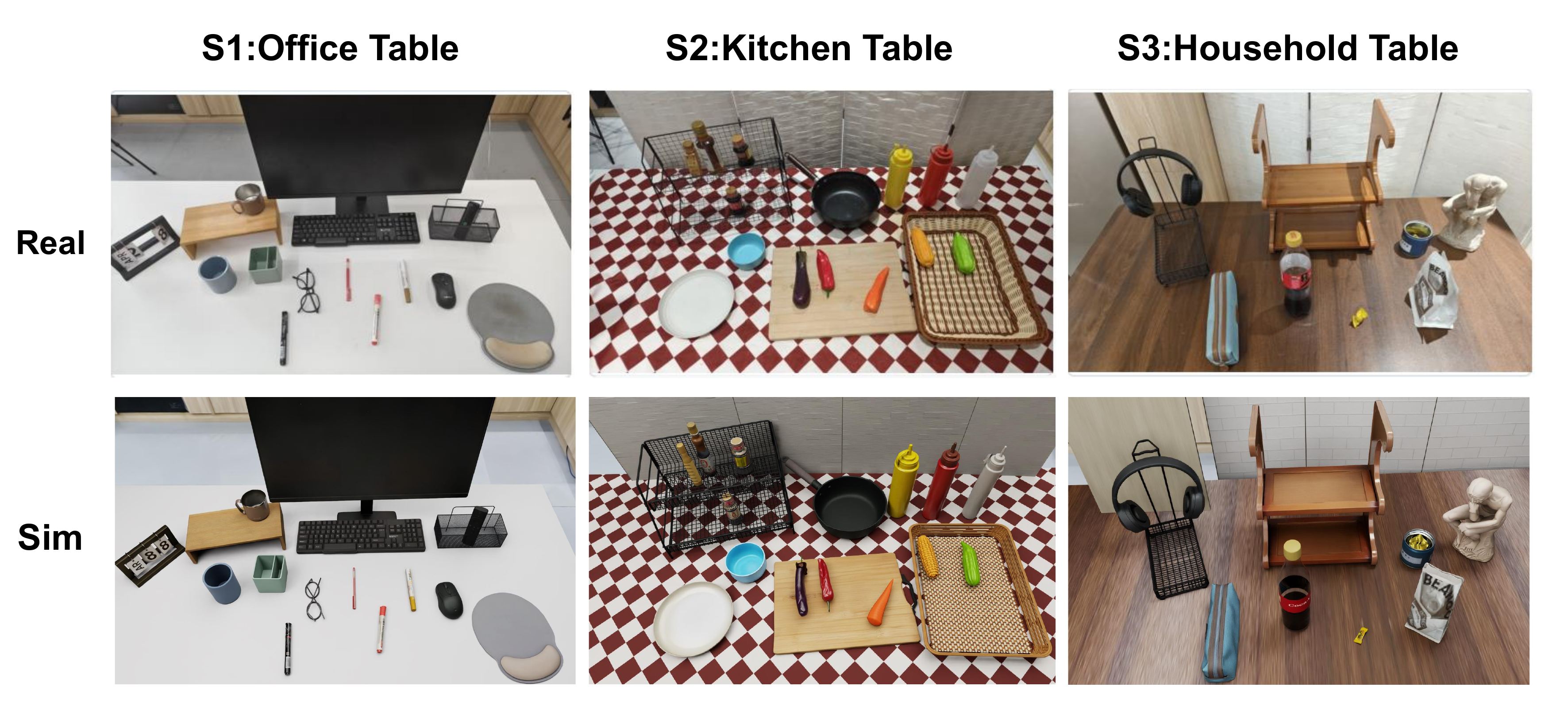}
\caption{\textbf{Paired physical and simulated experimental settings.}
Columns from left to right show Office Table (S1), Kitchen Table (S2), and Household Table (S3).
The top row shows the physical workspaces, and the bottom row shows their corresponding simulated environments.
Each column presents the same scene in the real world and simulation.}
\label{fig:environments}
\end{figure}

\paragraph{Experimental settings.}
We use a 14-DoF dual-arm ALOHA~\citep{zhao2023learning}
with two wrist cameras and one head camera, matching the simulation observations. The paired workspaces comprise Office Table (S1, six subtasks), Kitchen Table (S2, four) and Household Table (S3, four) (Figure~\ref{fig:environments}).
Each training setting uses one policy per scene.
R0 is initialized from the pretrained $\pi_{0.5}$ checkpoint~\citep{intelligence2025pi_}.
Each subsequent RSI round continues training from the preceding round's
policy using the accumulated dataset.
The value model uses the vision--language backbone initialized from the same checkpoint, augmented with a value prediction head. We build our agent system based on GPT-5.5 and use Isaac Sim as the simulation platform.
Subtasks share simulation/physical goal definitions (Table~\ref{si-tab:task_definitions}) and are evaluated individually, not as complete tidying sequences.

\paragraph{Training budgets and evaluation.}
R0/R1/R2 use 50/200/400 cumulative \emph{retained simulated training trajectories per subtask}. R0 is the seed policy; R1 and R2 follow collection-and-training updates.
+10 adds ten real-world human-in-the-loop~(HIL) refinement trajectories per subtask.
All tests are policy-only, without external recovery. We run ten physical trials per task and setting; simulation test counts vary by task--checkpoint pair. Scene-balanced means weight scenes equally.
Appendix~\ref{si-sec:protocol} provides allocation and statistical details.

\subsection{Does the Policy Improve Across RSI Rounds?}
\label{sec:sim_evolution}

\begin{figure}[t]
\centering
\includegraphics[width=\linewidth]{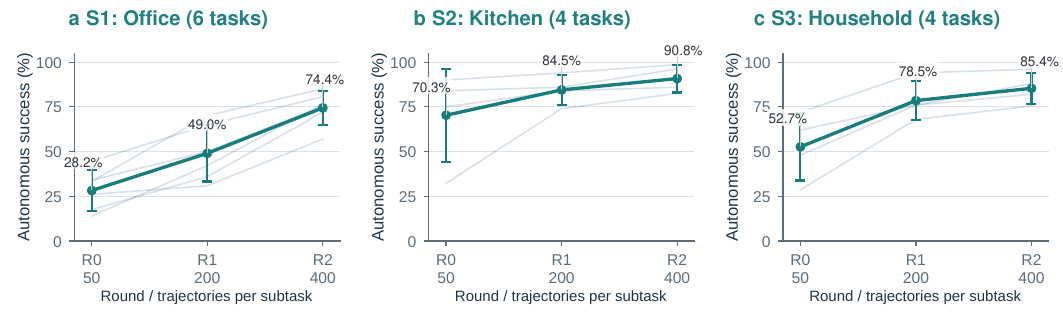}
\caption{\textbf{Autonomous policy performance across RSI rounds.}
Thin lines show individual subtasks; markers and error bars indicate scene means $\pm$ sample SD across subtasks.
Discrete rounds R0/R1/R2 use 50/200/400 retained simulated training trajectories per subtask, respectively.}
\label{fig:sim_evolution}
\end{figure}

Scene-balanced autonomous simulation success increases from 50.4\% at R0 to 70.7\% at R1 and 83.5\% at R2 (Figure~\ref{fig:sim_evolution}).
All 14 subtasks improve from R0 to R2, with substantial gains on initially challenging tasks.
For example, black-pen placement in S1 improves from 14.0\% to 76.5\%, whereas carrot placement in S2 starts near ceiling, leaving less room for improvement.
These results demonstrate that  EmbodiRSI's evaluated RSI rounds yield progressively stronger autonomous policies.

Qualitative observations are consistent with  EmbodiRSI's two feedback paths.
Early policies often continued nominal transport after missed grasps, whereas later policies grasped more reliably and sometimes reattempted unsuccessful interactions, with some reattempts leading to task completion.
Missed grasps occurred at peripheral object positions, although these positions were not shown to be unreachable or outside the training distribution.
Placement failures also occurred after successful grasping, particularly when objects missed narrow target openings.
These execution errors are distinct from task-selection errors. At the 50-demonstration initialization, the policy occasionally confused the red-cap-marker and red-pen instructions in S1.
Such confusion was not a recurring failure mode at larger simulation budgets, where most errors arose during execution of the requested manipulation.
Further details are provided in Appendix~\ref{si-sec:qualitative}.

\subsection{Does Feedback Help at Matched Trajectory Counts?}
\label{sec:ablation}
We compare  EmbodiRSI with Static collection in S1 and S3, the two more difficult scenes at initialization.
Both settings start from the same 50 demonstrations per subtask and use the same policy architecture and training recipe.
Static continues generating demonstrations under the original configuration distribution;  EmbodiRSI adds rollout-derived experience and targeted recollection.
Cumulative retained counts are matched at 200 and 400 trajectories per subtask.

\begin{table}[t]
\centering
\caption{\textbf{Effect of RSI under equal data budgets.}
Scene-balanced autonomous success (\%) across S1 and S3 at 400 simulated trajectories per subtask. Complete 200/400 comparisons, scene-level dispersion and task-level exceptions are retained in Appendix~\ref{si-sec:detailed_results}.}
\label{tab:ablation}
\small\renewcommand{\arraystretch}{1.15}
\begin{tabular}{@{}lccc@{}}
\toprule
\textbf{Method}
& \textbf{Simulation}
& \textbf{R2S2R}
& \textbf{R2S2R + Human in the Loop} \\
\midrule
EmbodiRSI w/o RSI
& 59.3 & 46.3 & 67.9 \\

EmbodiRSI
& \textbf{79.9} & \textbf{59.2} & \textbf{82.1} \\
\midrule
Gain (pp)
& +20.6 & +12.9 & +14.2 \\
\bottomrule
\end{tabular}

\end{table}

At a matched budget of 400 simulated trajectories per subtask,  EmbodiRSI's adaptive collection achieves 79.9\% scene-balanced simulation success, compared with 59.3\% for Static---a gain of 20.6 percentage points (Table~\ref{tab:ablation}). Adaptive collection yields higher scene-level mean success in both scenes at both evaluated budgets. This advantage extends to real-world execution, reaching 12.9 percentage points before physical refinement and 14.2 points after both policies are refined with ten human-in-the-loop~(HIL) trajectories per subtask. The benefits of adaptive simulation collection therefore persist after limited real-world refinement. Detailed task-level simulation results are provided in Appendix~\ref{si-sec:detailed_results}.

\subsection{Does RSI Enable Data-Efficient Real-World Adaptation?}
\label{sec:real_evaluation}
With 400 simulated and ten real-world refinement trajectories per subtask,  EmbodiRSI reaches 83.1\% scene-balanced autonomous real-world success, compared with 75.0\% for Real-only adaptation using 200 real demonstrations per subtask (Table~\ref{tab:real_main}).
Its observed mean is higher in each scene, using one-twentieth as many retained physical training trajectories.

\begin{table}[t]
\centering
\caption{\textbf{Data-efficient real-world adaptation.}
Success rates (\%) are mean $\pm$ sample SD across subtasks; the final column weights scenes equally.
R1/R2 use 200/400 simulated trajectories per subtask; +10 adds ten real human-in-the-loop~(HIL) trajectories.
Real-only labels give demonstration counts per subtask.
Each subtask/setting has ten autonomous test trials.}
\label{tab:real_main}
\small\setlength{\tabcolsep}{4pt}\renewcommand{\arraystretch}{1.15}
\begin{tabular}{@{}lcccc@{}}
\toprule
\textbf{Training setting} & \makecell{\textbf{S1: Office}\\$n=6$ tasks} & \makecell{\textbf{S2: Kitchen}\\$n=4$ tasks} & \makecell{\textbf{S3: Household}\\$n=4$ tasks} & \makecell[c]{\textbf{Scene-balanced}\\\textbf{mean}} \\
\midrule
R1 & $43.3 \pm 22.5$ & $60.0 \pm 8.2$ & $52.5 \pm 9.6$ & $51.9$ \\
R1+10 & $63.3 \pm 13.7$ & $82.5 \pm 5.0$ & $75.0 \pm 17.3$ & $73.6$ \\
R2 & $53.3 \pm 32.0$ & $70.0 \pm 14.1$ & $65.0 \pm 12.9$ & $62.8$ \\
R2+10 & $\textbf{81.7} \pm \textbf{13.3}$ & $\textbf{85.0} \pm \textbf{12.9}$ & $\textbf{82.5} \pm \textbf{9.6}$ & $\textbf{83.1}$ \\
\midrule
Real-only 10 & $13.3 \pm 13.7$ & $45.0 \pm 5.8$ & $20.0 \pm 8.2$ & $26.1$ \\
Real-only 100 & $56.7 \pm 15.1$ & $62.5 \pm 5.0$ & $52.5 \pm 5.0$ & $57.2$ \\
Real-only 200 & $70.0 \pm 6.3$ & $77.5 \pm 9.6$ & $77.5 \pm 9.6$ & $75.0$ \\
\bottomrule
\end{tabular}

\end{table}

In zero-shot deployment, both Static and RSI policies generally approached
the appropriate object despite pose variation. Narrow placement regions made
near-contact errors conspicuous, including pens released outside their holders.
Broader target regions were more forgiving, but constrained-placement trials
did not invariably fail. RSI policies showed more consistent grasp alignment
and sometimes completed an episode after a new attempt.

Before refinement, R2 already achieves 62.8\% physical success. Here
\emph{zero-shot} means no task-specific real-world policy refinement.
The policy generally approaches the instructed object, while
remaining failures concentrate near grasping and placement.
human-in-the-loop~(HIL) corrections address these residual errors rather than providing the entire strategy from scratch.
For example, S1 red-pen placement remains at 2/10 before refinement at
both simulation checkpoints but reaches 7/10 after R2 refinement; glasses placement already achieves 10/10 before refinement.

These patterns suggest complementary roles: simulation RSI develops
scene-specific capability through repeated practice and corrective experience, while physical refinement addresses residual deployment errors.
Figure~\ref{fig:real_examples} contrasts some real-world executions; Appendix~\ref{si-sec:detailed_results} retains all task-level outcomes and sampling intervals.
Reattempts during these tests are actions of the learned policy, not training-time agent recovery.
Further details are provided in Appendix~\ref{si-sec:detailed_results}.

\section{Limitations}
\label{sec:discussion}

Within the available time and computational budget, we prioritized evaluating scene-specific adaptation across the selected workspaces.
We plan to extend evaluation to held-out workspaces and longer improvement runs to assess generalization and sustained policy improvement.

Our evaluation centers on the complete pipeline, in which recovery, recollection and agent reasoning operate jointly.
While component-wise attribution is not the primary focus of this system-level study, we plan targeted ablations to better understand their contributions and interactions, informing further refinement of the pipeline.

\section{Conclusion}

EmbodiRSI enables recursive policy self-improvement within a real-to-sim-to-real (R2S2R) workflow by making experience acquisition responsive to the evolving policy. With agent assistance, Collaborative Error Correction provides corrective continuations from policy-failed states, while Adaptive Data Collection directs practice toward persistent weaknesses. Supported by task-specific simulation reconstructed from real-world observations and validated trajectory generation, these mechanisms turn rollout feedback into training experience for successive policy updates. After recursive improvement in simulation, the policy is refined with limited real-world data for autonomous deployment in the corresponding physical workspace. Experiments demonstrate improved autonomous performance across the evaluated rounds, gains over static collection at matched retained-trajectory counts, and data-efficient real-world adaptation. Together, these results highlight the value of refining not only the policy, but also the experience that drives its improvement.

\label{page:main_end}

\FloatBarrier
\phantomsection
\label{page:statements_start}
\section*{AI Use Statement}
Generative AI assisted earlier manuscript restructuring, language revision, literature checks and descriptive analysis code. Additional AI-assisted editing helped polish this manuscript,
including language revision, interpretation of existing results, LaTeX formatting, figure-layout exploration and editable vector diagrams. Submitted experimental images are retained from the supplied materials; generated layout concepts are not experimental evidence.
Manuscript preparation introduced no new robot trials or simulation
measurements. The diagrams are method illustrations, not generated experimental evidence. The reconstruction gallery includes labeled AI-generated scene inputs for illustration, not evaluation. The human authors retain responsibility for the final text, claims, references, implementation details and AI-use disclosure.

\section*{Reproducibility Statement}
The full code for the entire pipeline will be released soon.

\bibliographystyle{plainnat}
\bibliography{references}

\clearpage
\appendix
\numberwithin{equation}{section}
\numberwithin{figure}{section}
\numberwithin{table}{section}
\section{RSI implementation and agent coordination}
\label{si-sec:supervision}
This appendix expands the RSI loop in Section~\ref{sec:feedback}. It documents
state-conditioned handover, failure-attribution records and supervisory
coordination, distinguishing proposed operations, executed corrections and
final outcomes. Environment and trajectory construction are detailed separately
in Appendix~\ref{si-sec:infrastructure}.

\subsection{Task interfaces and recovery handover}
\label{si-sec:implementation_interfaces}
Task configurations associate the manipulated object and target anchor with a language instruction, grasp candidates, an ordered robot program and reset groups. Each associated waypoint follows the transformation of its source or target object. Members of a reset group share a sampled transformation, while sequential placement and collision handling can introduce dependencies between groups. These associations allow the same task-relative program to be re-grounded after a reset without specifying a new task goal.

The object-route generator returns intermediate world-frame poses with stage labels for lift, optional bypass and pre-placement. Its structured-output checks test object identity, numerical validity, pose representation and stage ordering before geometric review. Failed checks provide feedback for candidate revision. This separates a proposed route from its validation and subsequent execution; a plausible textual proposal alone is not an executable demonstration.

During monitored simulation collection, visual diagnosis and recovery planning are synchronous: the control loop does not advance the simulation while waiting for these operations. Recovery uses the current object and robot states. On return from a correction, observations are refreshed and completion is checked again; the continuation path can return control to the policy under the configured recovery gates. Reaching a feasible entry pose does not guarantee completion of the remaining manipulation.

Event records distinguish a recovery request, any corrective actions actually executed, and the final episode outcome. A request can fail before any correction is executed, while a partially executed correction can be followed by policy completion. Thus, triggering recovery, executing recovery and passing the final task check are distinct events. These training-collection events are also distinct from the policy-generated reattempts observed during unassisted evaluation.

% \FloatBarrier

% \subsection{Recovery experience within the RSI loop}
% Algorithm~\ref{alg:evolution} in Section~\ref{sec:method} summarizes the learning loop. Figure~\ref{si-fig:recovery_example} illustrates its state-conditioned recovery step during training-data collection, distinct from policy-generated reattempts during testing.
% \begin{figure}[!htbp]
% \centering
% \includegraphics[width=\linewidth]{figures/recovery_example.pdf}
% \caption{\textbf{Corrective experience from a simulated intermediate state.} The frames show an imperfect state, motion replanning and corrective execution during training-data collection. Externally generated recovery is disabled in performance tests; policy-generated reattempts remain part of autonomous execution.}
% \label{si-fig:recovery_example}
% \end{figure}

\FloatBarrier
\subsection{Grid attribution records and counting}
\label{si-sec:grid_attribution}

The grid analysis uses structured rollout records, including reset positions,
final outcomes and subject-attribution labels. A VLM request can assign a
failure to the source-object reset, target-anchor reset or neither; the request
uses the recorded execution evidence rather than requiring a new replay of the
rollout video. Existing attribution records can be reused, so issuing a new
VLM request is optional and does not replace the normalized-grid procedure.
For each subject and cell, valid completed episodes contribute to the
denominator, whereas only final failures with supported attribution to that
subject contribute to the numerator. Low-confidence and unassigned failures do
not increase the corresponding attributed-failure count. An episode that
finishes successfully after recovery is not counted as a final failure. 
% These statistics describe recorded outcomes, not a counterfactual estimate of whether an assisted episode would have succeeded without intervention.

\needspace{155pt}
\subsection{Supervisor scheduling and output validation}
\label{si-sec:supervisor_runtime}

\paragraph{Skills and workflow state.}
The supervisor coordinates the workflow through a typed skill library,
\begin{equation}
 \mathcal K=\{k^{\mathrm{scene}},k^{\mathrm{capture}},k^{\mathrm{grasp}},
 k^{\mathrm{config}},k^{\mathrm{motion}},k^{\mathrm{train}},k^{\mathrm{eval}}\},
\end{equation}
covering scene construction, observation capture, grasp inference and
verification, task-configuration generation, motion synthesis, policy training
and monitored evaluation. Each skill specifies typed inputs, explicit
preconditions, an execution backend and the outputs needed by downstream
operations. The persistent workflow state is
\begin{equation}
 s_t=(g,b,r,\mathbf z_t,\mathcal J_t,\mathcal A_t),
\end{equation}
where $g$ is the user-specified objective, $b$ the requested stopping boundary,
$r$ the scene and run identifier, $\mathbf z_t$ the stage states,
$\mathcal J_t$ the asynchronous jobs, and $\mathcal A_t$ the indexed artifacts.
Before execution, the supervisor constructs the dependency chain, assigns
stage-specific computing resources and fixes the requested stopping boundary.

\paragraph{Eligibility and stage completion.}
At time $t$, eligible skills satisfy
\begin{equation}
 \mathcal K_t^{\mathrm{ready}}=
 \{k_i\in\mathcal K\mid\mathrm{Pre}_i(s_t)=1,
 \ \mathrm{Valid}(\mathcal A_i^{\mathrm{req}})=1,\ k_i\preceq b\}.
\end{equation}
Here $\mathrm{Pre}_i$ represents dependency and resource constraints,
$\mathcal A_i^{\mathrm{req}}$ is the required upstream artifact set, and
$k_i\preceq b$ prevents execution beyond the requested boundary. The supervisor
selects $k_t=\pi_{\mathrm{sup}}(s_t,\mathcal K_t^{\mathrm{ready}})$. Independent
scene- or object-level jobs can be dispatched in parallel on different devices.
Long-running skills return a durable job handle and are monitored through their
runtime state, logs and artifact manifests. Successful completion requires both
a normal process exit and valid required outputs:
\begin{equation}
 \mathrm{Complete}(k_i)=\mathbb I[e_i=0]
 \prod_{a\in\mathcal A_i^{\mathrm{out}}}\mathbb I[\mathrm{Valid}(a)=1].
\end{equation}
Here $e_i$ is the exit code and $\mathcal A_i^{\mathrm{out}}$ contains outputs
required downstream. Persistent services use readiness checks rather than
process-exit semantics.

\paragraph{Failure handling.}
When a precondition or artifact check fails, the supervisor localizes the
earliest invalid dependency and requests the corresponding upstream repair,
rather than restarting the entire workflow. Transient communication failures
are handled by recovering the latest durable job and stage state before making
a new scheduling decision. These runtime checks support the scene-to-policy
coordination described in Section~\ref{sec:method} without treating a process exit or an
accepted tool request as sufficient evidence of task completion.

\FloatBarrier

\section{Environment and trajectory infrastructure}
\label{si-sec:infrastructure}
The simulator and trajectory generator provide the reusable infrastructure for
RSI. This appendix retains the scene compilation, collision rectification,
trajectory validation and reset-aware execution details summarized in
Sections~\ref{sec:scene_generation} and~\ref{sec:trajectory_main}. These tools
serve initial collection, state-conditioned recovery and targeted recollection;
they are not separate deployment-time controllers.

\subsection{Agentic Real2Sim Scene Construction and Refinement}

\label{si-sec:scene_details}
\begin{figure}[t]
\centering
\includegraphics[width=\linewidth]{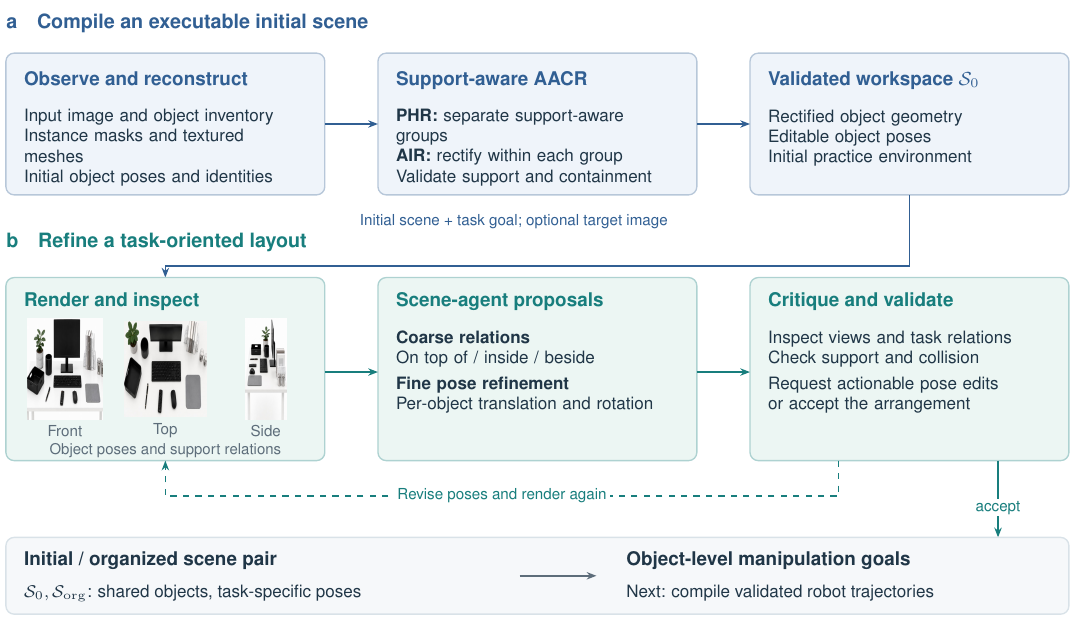}
\caption{\textbf{From deployment observations to task-defining scenes.}
Object reconstruction and support-aware collision rectification produce the initial scene.
The scene agent iteratively refines object poses using multi-view feedback and geometric validation to obtain the target scene.
The scene pair defines manipulation goals for trajectory generation.}
\label{fig:agent_scene}
\end{figure}

A reconstructed workspace is useful for learning only when it supports both
physically valid interactions and explicit task goals. We therefore formulate
Real2Sim construction as an \emph{agentic scene compiler}: visual and semantic
reasoning specifies the objects and their intended relations, while geometric
tools implement and check the corresponding scene operations. Given a tabletop
image, optionally accompanied by a target image and language prompt, the compiler
produces an initial scene and an organized target layout
(Figure~\ref{fig:agent_scene}). Their difference defines the object-level goals
for demonstration generation.

\paragraph{Object-centric reconstruction.}
A vision--language model (VLM) identifies the table and visible movable objects
in the input image $I_0$. SAM~3 produces instance masks and SAM~3D reconstructs
the corresponding textured meshes~\citep{carion2026sam,chen2026sam}. The scene is
represented as
\begin{equation}
 \mathcal S_0=\{o_T,o_1,\ldots,o_N\},\qquad
 o_i=(c_i,m_i,T_i^0),
 \label{eq:scene_repr}
\end{equation}
where $o_T$ is the table, and $c_i$, $m_i$ and $T_i^0\in\mathrm{SE}(3)$ are an
object's semantic category, mesh and initial pose. Object identity is maintained
through scene editing and trajectory generation, so that an instruction refers
to the same object in the rendered workspace, geometric checks and robot program.

\paragraph{Simulation instantiation.}
The scene agent configures the reconstructed assets for simulation,
setting and adjusting object scale, poses, and physical properties to
obtain an executable task environment.
The resulting scene undergoes collision and support checks before
task and trajectory generation.

\paragraph{Support-aware collision rectification.}
Imperfect reconstruction can leave objects interpenetrating, particularly in
cluttered tabletop scenes. Agent-Guided Adaptive Collision Rectification (AACR)
resolves these collisions while preserving the support and containment relations
inferred from $I_0$. The agent first partitions the non-table objects into
support-aware groups $\mathcal G=\{G_1,\ldots,G_K\}$.
\emph{Parallel Horizontal Rectification} separates intersecting groups by
selecting a central root and moving other groups outward. A group moves as a
unit, preserving its internal relations; only groups with disjoint swept
regions are adjusted in the same parallel batch. The settled set is then
updated before the next batch is processed.

\emph{Adaptive Intra-group Rectification} resolves the remaining collisions
within each group. The agent selects a correction according to the object
relation: horizontal separation for objects sharing a support, vertical
adjustment for containment, or bounded tilt for insertion-like arrangements.
The selected geometric operator applies and validates the pose change. Thus,
semantic reasoning determines which adjustment is appropriate, rather than
replacing collision checking. The group-selection and displacement rules are
given in Section~\ref{si-sec:geometry} below.

\paragraph{Task-oriented scene organization.}
The scene agent compares the original image $I_0$ with a rendering of the
rectified scene and generates an organization instruction $\mathcal L$.
A supplied target image or language prompt can further specify the intended
layout. The agent first uses relation-specific tools, such as
\texttt{on-top-of}, \texttt{inside} and \texttt{beside}, to establish a coarse
arrangement. It then refines object translations and rotations, with individual
object agents operating in parallel. Rendered views and structured object states
make the consequences of an edit available for critique and revision. This
coarse-to-fine process separates deciding \emph{which relation should hold} from
finding poses that realize that relation without invalid geometry.

The accepted target scene is
\begin{equation}
 \mathcal S_{\mathrm{org}}
 =\{(c_i,m_i,T_i^{\mathrm{org}})\}_{i\in\{T,1,\ldots,N\}}.
 \label{eq:organized_scene}
\end{equation}
The initial and organized layouts specify what the robot should change.
During task compilation, these goals are associated with language instructions,
grasps, waypoints and reset groups. Objects in a reset group share a sampled
transformation. Source- and target-associated waypoints subsequently follow
their respective anchors, allowing the same task to be practiced under different
configurations without redefining its goal.

\subsection{Agentic Trajectory Generation}

\label{si-sec:trajectory_details}
The scene pair defines the desired transition, but not a feasible way for the
robot to execute it. The trajectory-generation agent
$\mathcal A_{\mathrm{traj}}$ bridges this gap by proposing object motions,
projecting them through candidate grasps and validating the resulting robot
programs. Its interface is
\begin{equation}
 \mathcal A_{\mathrm{traj}}:
 (\mathcal S_0,\mathcal S_{\mathrm{org}},l)
 \longmapsto(\mathcal C,\mathcal D_{\mathrm{demo}},\mathcal E),
 \label{eq:trajectory_interface}
\end{equation}
where $l$ is the task instruction, $\mathcal C$ is a library of executable
configurations, $\mathcal D_{\mathrm{demo}}$ contains accepted demonstrations
and $\mathcal E$ records validation evidence. The same trajectory tools later
support recovery from policy-failed states and recollection under revised
reset ranges. Demonstration generation is therefore a reusable part of the
learning loop, not a one-time preprocessing stage.

\paragraph{Object-centric task compilation.}
The agent compares $\mathcal S_0$ and $\mathcal S_{\mathrm{org}}$, identifies
objects whose states must change and compiles an ordered list of subtasks,
\begin{equation}
 \mathcal T=(\chi_j)_{j=1}^{M},\qquad
 \chi_j=(o_{i_j},b_j,\rho_j).
 \label{eq:task_compilation}
\end{equation}
Here $o_{i_j}$ is the manipulated object, $b_j$ its target anchor or support,
and $\rho_j$ the desired relation. The order accounts for accessibility,
target occupancy and earlier placements; the subtask count $M$ need not equal
the number of reconstructed objects $N$. For each subtask, a VLM planner
proposes object-space routes comprising lift, optional bypass and pre-placement
poses. These routes are simplified, refined for clearance and densified to
bound translation and rotation between consecutive states. Planning in object
space allows the intended motion to be evaluated before a particular grasp is
chosen.

\paragraph{Grasp-conditioned motion synthesis.}
GraspGen supplies candidate grasps~\citep{murali2025graspgen}, which are screened
for grasp quality, endpoint reachability and collision clearance. For a grasp
$T_{g,i}^0$ on object $i$, the object-relative gripper transform is preserved
along an object route:
\begin{equation}
 T_{o_i\rightarrow g}=(T_{o_i}^0)^{-1}T_{g,i}^0,
 \qquad T_{g,i}^{j}=T_{o_i}^{j}T_{o_i\rightarrow g}.
 \label{eq:grasp_projection}
\end{equation}
An accepted route can thus be reused across grasp candidates without
regenerating its semantic path. Pick and place are checked jointly. When an endpoint pair is unreachable, pick and place poses are repaired
jointly using bounded $\mathrm{SE}(3)$ adjustments.
The repaired poses must satisfy the shared object-relative grasp relation
in Eq.~\eqref{eq:grasp_projection}. Pre-grasp, gripper closure, release and post-release retract
complete the waypoint program.

\paragraph{Validation-guided repair.}
A plausible object route does not by itself establish a valid demonstration.
EmbodiRSI uses three checks before admitting a configuration to $\mathcal C$.
First, a swept full-mesh test checks the carried object along each interpolated
segment, with narrowly scoped exemptions for intended support and target
contact. Second, multi-view renderings allow a VLM reviewer to assess route
efficiency, orientation consistency and target approach. Third, cuRobo validates
the complete robot program sequentially, initializing each transition from the
preceding terminal joint state~\citep{sundaralingam2026curobov2}. Because the
carried object's swept geometry is checked separately, its stale source-pose
mesh is disabled after grasping; robot collision checking against the rest of
the scene remains active.

When a transition is infeasible, the agent first attempts route contraction,
then searches bounded local pose adjustments around the first infeasible
waypoint. A repaired candidate must pass the geometric and sequential-motion
checks again, together with renewed visual review when its object route
changes. Only programs feasible through post-release retract are accepted.
These checks connect high-level proposals to executable supervision while
allowing invalid candidates to be repaired rather than treated as training data.

\paragraph{Reset-aware execution and data curation.}
At collection time, scene objects are randomized within their admissible reset
ranges. Once physics has settled, each stored waypoint is re-grounded to the
measured pose of its associated source or target anchor $r(j)$:
\begin{equation}
 \widetilde T_{ee}^{j}
 =T_{r(j)}^{\mathrm{cur}}(\overline T_{r(j)})^{-1}\overline T_{ee}^{j},
 \label{eq:reset_regrounding}
\end{equation}
where overbars denote poses stored in the configuration. Each segment is
replanned from the measured joint state and then executed. Transit motions
permit broader local pose search, whereas grasp and placement use tightly
bounded planar fallbacks to preserve the task relation.

Execution supplies a fourth validation layer. The system checks the
object--gripper relation after grasping, verifies transport at designated
waypoints and compares the settled object pose after release with the
reset-adjusted target. A generated demonstration is retained only when
\begin{equation}
 \operatorname{Accept}(\tau)
 =V_{\mathrm{mesh}}\land V_{\mathrm{view}}
 \land V_{\mathrm{motion}}\land V_{\mathrm{exec}}.
 \label{eq:accept}
\end{equation}
The terms denote swept-mesh, visual-review, sequential-motion and execution-time validity.
Failed demonstrations are rejected and resampled. Accepted records contain multi-camera observations, joint states, velocities, efforts and commanded actions in LeRobot format. Together, proposal, validation, repair and execution turn an object-level task into varied training demonstrations without manual waypoint scripting.

\subsection{Scene geometry}
\label{si-sec:geometry}

\paragraph{Parallel Horizontal Rectification.}
AACR partitions non-table objects into support-aware groups
$\mathcal G=\{G_1,\ldots,G_K\}$. PHR treats each group as a unit while resolving
inter-group overlap. Let $\mathbf p_k$ denote the center of $G_k$ projected onto
the tabletop plane, represented in world coordinates with zero vertical
component. The root minimizes the sum of planar distances to the other groups;
each non-root group is assigned the corresponding outward direction:
\begin{equation}
 r=\arg\min_k\sum_{j\ne k}\|\mathbf p_k-\mathbf p_j\|_2,\qquad
 \mathbf d_k=\frac{\mathbf p_k-\mathbf p_r}{\|\mathbf p_k-\mathbf p_r\|_2},
 \quad k\ne r,\ \mathbf p_k\ne\mathbf p_r.
 \label{si-eq:phr_root}
\end{equation}
The displacement $\delta\mathbf d_k$ is horizontal. Starting from the settled
anchor set $\mathcal H_0=\{G_r\}$, each unresolved group is assigned the minimum
displacement that removes overlap with the settled groups:
\begin{equation}
 \delta_k^*=\min\{\delta\ge0\mid(G_k+\delta\mathbf d_k)\cap\mathcal H_t=\emptyset\},
 \qquad \widehat G_k=G_k+\delta_k^*\mathbf d_k.
 \label{si-eq:phr_shift}
\end{equation}
Intersections refer to occupied geometry. A conflict-aware parallel batch
$\mathcal B_t$ contains only groups whose proposed swept regions are disjoint:
\begin{equation}
 \operatorname{Sweep}(G_i,\widehat G_i)\cap
 \operatorname{Sweep}(G_j,\widehat G_j)=\emptyset,
 \qquad G_i,G_j\in\mathcal B_t,\ i\ne j.
 \label{si-eq:phr_batch}
\end{equation}
Here $\operatorname{Sweep}$ denotes the spatial region traversed during the
proposed translation. The selected groups are moved simultaneously, and the
settled set is updated as
\begin{equation}
 \mathcal H_{t+1}=\mathcal H_t\cup
 \{\widehat G_k\mid G_k\in\mathcal B_t\}.
 \label{si-eq:phr_update}
\end{equation}
The equations specify the non-degenerate geometry. AIR subsequently addresses
within-group overlap using the relation-specific horizontal, vertical and
bounded-tilt corrections described in Section~\ref{si-sec:scene_details}.

\subsection{Sequential motion execution}
\label{si-sec:sequential_execution}

Robot transitions are planned sequentially, with the terminal joint state of
one transition used to initialize the next:
\begin{equation}
 \xi^*_{i,k}=\operatorname{Plan}(q_{i,k-1},T^k_{g,i},\mathcal W_{i,k}),\qquad
 q_{i,k}=\operatorname{Last}(\xi^*_{i,k}).
 \label{si-eq:sequential_planning}
\end{equation}
Here $\mathcal W_{i,k}$ is the collision world used for the transition. At
collection time, Eq.~\eqref{eq:reset_regrounding} re-grounds each
waypoint using the current pose of its source object or target anchor. This
preserves task-relative motion after a reset. The corresponding geometry and
motion checks are repeated after a repair; visual review is repeated when the
object route changes.

\FloatBarrier

\section{Task environments and scene reconstruction results}
\label{si-sec:tasks}
\subsection{Robot platform and evaluated tasks}
Section~\ref{sec:experiments} and Figure~\ref{fig:environments} present the
robot platform and paired physical/simulated workspaces. Table~\ref{si-tab:task_definitions}
lists all 14 object-level task goals shared by each simulation--real pair.
One policy covers the subtasks of each scene and training setting; tasks are
evaluated separately. The matched-collection comparison uses S1 and S3.

\begin{table}[!htbp]
\centering
\caption{\textbf{Manipulation subtask definitions.} Initial and goal relations are kept fixed across the corresponding simulation and real-world task definitions.}
\label{si-tab:task_definitions}
\small\renewcommand{\arraystretch}{1.04}\setlength{\tabcolsep}{6pt}
\begin{tabularx}{\linewidth}{@{}lX@{}}
\toprule\textbf{Subtask}&\textbf{Goal}\\\midrule
S1-1&Place glasses on the stand.\\
S1-2&Put the red-cap marker in the green cup.\\
S1-3&Put the yellow-cap marker in the basket.\\
S1-4&Place the mouse on the mouse pad.\\
S1-5&Put the red pen in the green box.\\
S1-6&Put the black pen in the pen holder.\\\midrule
S2-1&Place the blue bowl on the plate.\\
S2-2&Put the carrot in the basket.\\
S2-3&Put the eggplant in the basket.\\
S2-4&Put the red chilli in the basket.\\\midrule
S3-1&Place the soda bottle on the shelf.\\
S3-2&Put the pencil case in the metal box.\\
S3-3&Place the snack bag on the shelf.\\
S3-4&Put the candy wrapper in the can.\\
\bottomrule\end{tabularx}
\end{table}

\FloatBarrier

\subsection{Reconstruction and organization examples}
\begin{figure}[!htbp]
\centering
\includegraphics[width=.85\linewidth]{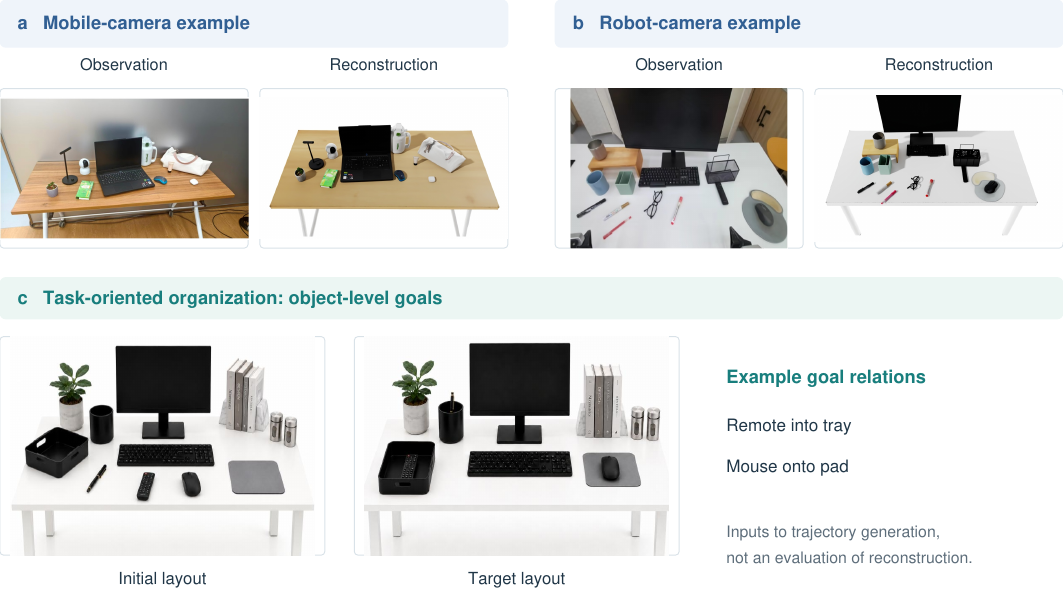}
\caption{\textbf{Qualitative examples of scene reconstruction and organization.}
\textbf{a,b}, Mobile-camera and robot-camera observations paired with reconstructed scenes; views are not pixel-aligned.
\textbf{c}, Initial and target layouts illustrating object-level manipulation goals.}
\label{si-fig:reconstruction_pairs}
\end{figure}

Figure~\ref{si-fig:reconstruction_pairs} places selected observations beside their reconstructions and illustrates how an initial layout defines an organization goal. The examples complement the actual evaluation environments in Figure~\ref{fig:environments}.

\begin{figure}[!htbp]
\centering
\includegraphics[width=.95\linewidth]{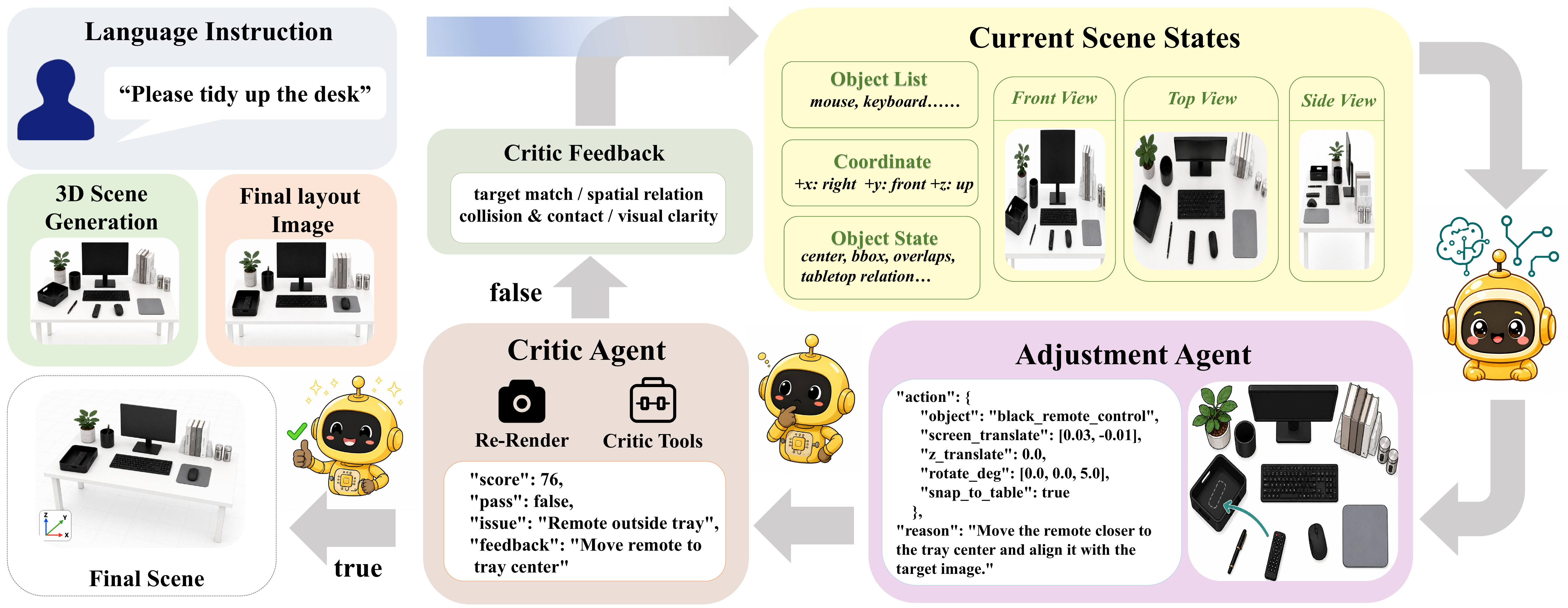}
\caption{\textbf{Iterative scene organization through critique, editing, and validation.}
The critic compares rendered views and object states with target relations to guide pose adjustments.
Collision and support checks validate each edit before re-rendering.
The accepted layout defines object-level goals for trajectory generation.}
\label{si-fig:scene_refinement}
\end{figure}

Figure~\ref{si-fig:scene_refinement} expands the scene-agent workflow using an example initial and target layout. It separates semantic assessment from deterministic validation and complements the method pipeline in Figure~\ref{fig:agent_scene}.
% \FloatBarrier

\begin{figure}[!htbp]
\centering
\includegraphics[width=.98\linewidth]{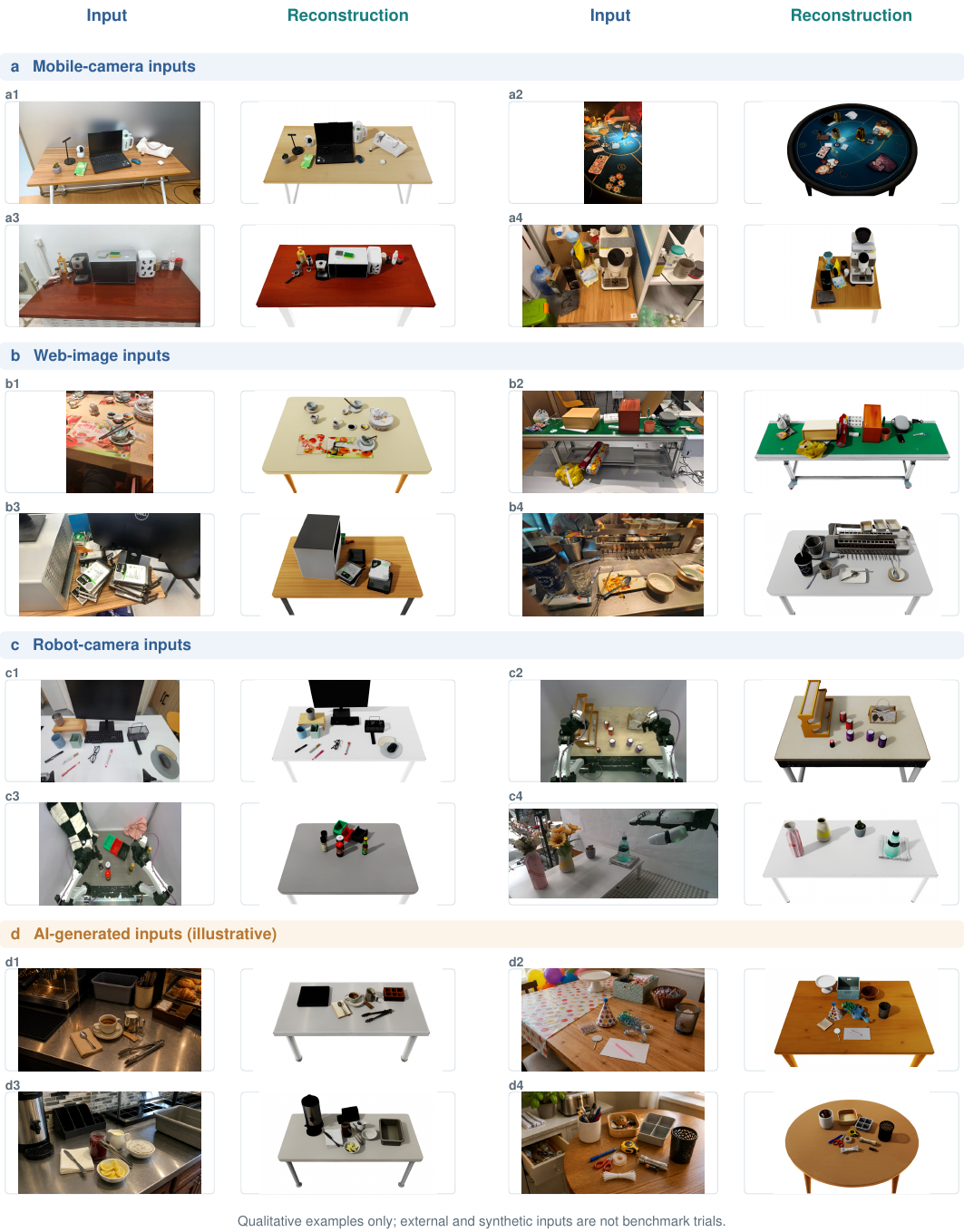}
\caption{\textbf{Qualitative input--reconstruction pairs.}
\textbf{a--d}, Mobile-camera, web, robot-camera, and AI-generated inputs, respectively.
Each pair shows the input on the left and reconstruction on the right.
Examples extend beyond the evaluation benchmark; views are not aligned in pose or scale.}
\label{si-fig:gallery}
\end{figure}

Figure~\ref{si-fig:gallery} presents additional input--reconstruction pairs.

% \FloatBarrier

\section{Training data and evaluation protocol}
\label{si-sec:protocol}
\subsection{Training data and retained-trajectory budgets}
Every stated data budget is the number of retained training trajectories for \emph{each subtask}. Data from all subtasks of a scene train one scene-specific policy. Table~\ref{si-tab:data_sources} separates the sources of training experience; Tables~\ref{si-tab:data_schedule} and~\ref{si-tab:scene_budgets} give per-subtask and scene-total counts.

\begin{table}[!htbp]
\centering
\caption{\textbf{Experience sources and retained-data notation.} The notation separates monitored training rollouts, corrective data and newly generated demonstrations. The reported quotas distinguish newly generated from rollout-derived data, without a separate count for each rollout-derived source. No numerical allocation between base and targeted collection is inferred from those quotas. Policy and correction symbols denote sources within episodes, not separately counted episode datasets.}
\label{si-tab:data_sources}
\small\renewcommand{\arraystretch}{1.16}\setlength{\tabcolsep}{5pt}
\begin{tabularx}{\linewidth}{@{}L{.20\linewidth}X L{.28\linewidth}@{}}
\toprule
\textbf{Data item} & \textbf{Source and learning role} & \textbf{Budget accounting} \\
\midrule
$\mathcal D_0$ & Initial validated simulation demonstrations. & 50 per subtask at R0. \\
$\mathcal D_k^{\mathrm{mon}}$ & Policy-generated trajectories recorded during monitored training collection. & Part of rollout-derived data. \\
$\mathcal D_k^{\mathrm{rec}}$ & Corrections generated from eligible intermediate states reached during collection. & Part of rollout-derived data. \\
$\mathcal D_k^{\mathrm{roll}}$ & Retained data assembled from monitored and corrective sources. & 50 added at R1; 100 at R2. \\
$\mathcal D_k^{\mathrm{new}}$ & Newly generated demonstrations after collection configurations are updated. & 100 added at each round. \\
\midrule
Real HIL data & Real-world training rollouts with operator corrections. & 10 per subtask; pooled within scene. \\
Real test records & Autonomous policy evaluation, independent of HIL training. & Ten trials per subtask and setting; not training data. \\
\bottomrule
\end{tabularx}

\end{table}

R0 uses 50 generated demonstrations per subtask. R1 adds 100 newly generated
and 50 rollout-derived trajectories; R2 adds 100 of each, giving cumulative
budgets of 200 and 400. The counts describe retained training sets, not total
acquisition attempts, transitions or computation.

\paragraph{Training-data selection.}
Our learning formulation supports joint training on successful and
unsuccessful trajectories. In the reported experiments, however, only
trajectories that ultimately completed the task were retained for policy
training. These include episodes that succeeded after corrective
intervention and may therefore contain intermediate policy errors.
Unsuccessful monitored rollouts informed failure attribution and targeted
collection but were excluded from the policy-training dataset.
All reported training budgets count retained successful trajectories.

\begin{table}[!htbp]
\centering
\caption{\textbf{Simulation training schedule per subtask.} The two data columns are additions at the indicated stage; the total is cumulative. For R1/R2, newly generated demonstrations are $\mathcal D_k^{\mathrm{new}}$ and retained rollout-derived data are $\mathcal D_k^{\mathrm{roll}}$. R0 contains the initial generated demonstrations. Counts do not include performance-test episodes.}
\label{si-tab:data_schedule}
\small\renewcommand{\arraystretch}{1.15}
\begin{tabular}{@{}lccc@{}}
\toprule
\textbf{Stage} & \makecell{\textbf{Newly generated}\\\textbf{demonstrations}} & \makecell{\textbf{Retained rollout-}\\\textbf{derived data}} & \makecell{\textbf{Cumulative}\\\textbf{total}} \\
\midrule
R0 & 50 & 0 & 50 \\
R1 & 100 & 50 & 200 \\
R2 & 100 & 100 & 400 \\
\bottomrule
\end{tabular}

\end{table}

\begin{table}[!htbp]
\centering
\caption{\textbf{Scene-level trajectory budgets.} Counts equal the reported per-subtask budget multiplied by the number of subtasks covered by the scene policy. Real-only 10 and HIL+10 contain the same number of trajectories but are different training settings. The last row is evaluation, not training; each real-world subtask is tested in ten autonomous rollouts.}
\label{si-tab:scene_budgets}
\small\renewcommand{\arraystretch}{1.15}
\begin{tabularx}{\linewidth}{@{}Xccc@{}}\toprule
\textbf{Dataset / evaluation}&\textbf{S1 (6 tasks)}&\textbf{S2 (4 tasks)}&\textbf{S3 (4 tasks)}\\\midrule
R0 simulation training&300&200&200\\
R1 simulation training&1,200&800&800\\
R2 simulation training&2,400&1,600&1,600\\
HIL+10 real training&60&40&40\\
Real-only 10 training&60&40&40\\
Real-only 100 training&600&400&400\\
Real-only 200 training&1,200&800&800\\\midrule
Real test episodes per setting&60&40&40\\\bottomrule
\end{tabularx}
\end{table}

\FloatBarrier
\subsection{Real-world refinement and comparison settings}
\label{si-sec:real_refinement}
\label{sec:real_refinement}
For each scene, the simulation-evolved policy is refined using 10 retained real-world HIL training trajectories per subtask. An operator corrects undesirable execution, such as slippage or inaccurate placement, and the retained data are annotated for the same conditioned training objective used in simulation. Data are pooled across subtasks to update the shared scene policy. Final real-world test rollouts are collected independently of HIL training.

The Real-only baselines use real-world demonstrations at budgets of 10, 100 and 200 trajectories per subtask, without the task-specific simulation stage. These demonstrations are collected by trained data-collection operators following a standard operating procedure. Static simulation collection continues from the initial demonstration dataset under the original reset distribution; EmbodiRSI instead combines rollout-derived experience with targeted recollection. Static and adaptive comparisons share the 50-trajectory-per-subtask initial dataset and match cumulative retained trajectory counts. The compared policies use the same architecture and training recipe. Table~\ref{si-tab:comparison_settings} summarizes these settings. Equal real-world trajectory budgets do not imply identical datasets; Real-only demonstrations and HIL corrections are distinct collection settings.

\begin{table}[!htbp]
\centering
\caption{\textbf{Training and collection settings.} Budgets are retained trajectories per subtask. The table distinguishes the adaptation data sources and configuration-sampling procedures. Equal real-data counts do not imply identical trajectories.}
\label{si-tab:comparison_settings}
\small\renewcommand{\arraystretch}{1.18}\setlength{\tabcolsep}{5pt}
\begin{tabularx}{\linewidth}{@{}L{.22\linewidth}>{\raggedright\arraybackslash}X>{\raggedright\arraybackslash}X>{\raggedright\arraybackslash}X@{}}
\toprule
\textbf{Setting} & \textbf{Real-only} & \textbf{Static simulation} & \textbf{EmbodiRSI (adaptive)} \\
\midrule
Scene policy & One per scene & One per scene & One per scene \\
Architecture & \multicolumn{3}{l}{Same policy architecture across compared groups} \\
Initial simulation data & Not used & Shared 50 demonstrations per subtask & Same initial dataset as Static \\
Subsequent reset sampling & Real collection & Original configuration distribution & Feedback-guided configuration update \\
Simulation trajectory budget & None & 200 or 400 & 200 (R1) or 400 (R2) \\
Training recipe & \multicolumn{3}{l}{Same policy-training recipe across compared groups} \\
Real training source & Demonstrations at 10/100/200 per subtask & Zero-shot or +10 HIL & Zero-shot or +10 HIL \\
Role in comparison & Physical-data reference & Static-collection control & Adaptive-collection workflow \\
\bottomrule
\end{tabularx}

\end{table}

\subsection{Autonomous evaluation protocol}
\label{si-sec:evaluation}
\label{sec:evaluation}
Each subtask is evaluated separately with the scene's trained policy. Agent takeover, planner-generated recovery and human correction are disabled; policy-generated reattempts remain part of autonomous execution. Outcomes are scored against the corresponding task definition. Here zero-shot real deployment means no task-specific real-world policy refinement; the simulation itself is constructed from observations of the target workspace.

In simulation, each task--checkpoint pair is tested in repeated rollouts under its evaluation reset configuration. The number of episodes varies across pairs. Success is $r_{s,j}=100N^{\mathrm{succ}}_{s,j}/N^{\mathrm{test}}_{s,j}$, where $N^{\mathrm{test}}_{s,j}$ is the total number of attempted evaluation episodes. These performance tests are distinct from monitored training collection, denoted by $\mathcal D^{\mathrm{mon}}$.

Real-world evaluation uses ten autonomous rollouts per subtask and setting, separate from HIL training collection, giving 140 test episodes for a fully evaluated setting. The empirical task rate is $r_{s,j}=100N^{\mathrm{succ}}_{s,j}/10$. Subtasks sharing a policy are not independently trained policy replicates.

\subsection{Statistical analysis and numerical precision}
\label{si-sec:statistics}
\label{sec:statistical_analysis}
Scene-level standard deviations describe variation across subtasks sharing a policy, not independent training runs. Task-matched comparisons do not imply episode-wise pairing of resets.
For scene $s$ with $M_s$ subtasks and task success percentages $r_{s,j}$, we report
\begin{equation}
 \overline r_s=\frac{1}{M_s}\sum_{j=1}^{M_s}r_{s,j},\qquad
 s_s=\sqrt{\frac{1}{M_s-1}\sum_{j=1}^{M_s}(r_{s,j}-\overline r_s)^2}.
\end{equation}
Here $M_1=6$ and $M_2=M_3=4$. Real-world rates use integer success counts; simulation summaries use the tabulated task-level percentages. For a fixed real-world task and trained policy, Figure~\ref{si-fig:task_gains} additionally reports pointwise 95\% Wilson intervals under an independent binomial model. For $\widehat p=N^{\mathrm{succ}}/n$, $n=10$ and $z_{.975}=1.959964$, the interval is
\begin{equation}
 \frac{\widehat p+z_{.975}^2/(2n)\ \pm\ z_{.975}\sqrt{\widehat p(1-\widehat p)/n+z_{.975}^2/(4n^2)}}{1+z_{.975}^2/n}.\label{si-eq:wilson}
\end{equation}

The main aggregate is the scene-balanced mean $\overline r=\frac{1}{3}\sum_{s=1}^{3}\overline r_s$, reported without an overall SD to keep it distinct from the within-scene statistics. Pooled real-world success counts are a different weighting: for example, R2+10 has 116/140 successes, whereas its scene-balanced mean is 83.1\%. Collection-method differences are formed per subtask before summarization, $d_{s,j}=r^{\mathrm{adaptive}}_{s,j}-r^{\mathrm{static}}_{s,j}$. Their mean and sample SD are reported in percentage points. Equal-scene collection-comparison means use S1 and S3 only.

All calculations retain the supplied input precision and use round-half-up to one decimal only for final displayed percentages, SDs and gains. Simulation collection comparisons use the complete S1/S3 results at both matched budgets; no Static result is extrapolated to S2. In task-level direction counts, increases, ties and decreases refer to observed success rates, not hypothesis tests.

\FloatBarrier

\section{Detailed results and behavioral observations}
\label{si-sec:detailed_results}
\label{si-sec:qualitative}
Task-level results and qualitative observations are grouped below; experimental
and statistical definitions are provided in Appendix~\ref{si-sec:protocol}.

\subsection{Simulation evolution}
Table~\ref{si-tab:sim_tasks} reports per-task simulation success across collection-and-training rounds.
\begin{table}[!htbp]
\centering
\caption{\textbf{Task-level simulation success (\%).}
Scene summaries report mean $\pm$ sample SD across subtasks (S1: six; S2/S3: four each), not confidence intervals.
The final row weights scenes equally and reports no SD.
Numbers in column headers denote simulated training trajectories per subtask.}
\label{si-tab:sim_tasks}
\small\setlength{\tabcolsep}{5pt}\renewcommand{\arraystretch}{1.13}
\begin{tabularx}{\linewidth}{@{}Xccc@{}}
\toprule
\textbf{Task} & R0 (50) & R1 (200) & R2 (400) \\
\midrule
\multicolumn{4}{@{}l}{\textbf{S1: Office Table}} \\
Glasses $\rightarrow$ stand & 44.5 & 64.9 & 80.7 \\
Red-cap marker $\rightarrow$ green cup & \ensuremath{34.0} & \ensuremath{50.1} & \ensuremath{75.3} \\
Yellow-cap marker $\rightarrow$ basket & 26.1 & \ensuremath{30.9} & \ensuremath{57.0} \\
Mouse $\rightarrow$ mouse pad & 33.3 & 70.1 & 85.1 \\
Red pen $\rightarrow$ green box & 17.5 & \ensuremath{36.0} & \ensuremath{72.0} \\
Black pen $\rightarrow$ pen holder & 14.0 & \ensuremath{42.2} & \ensuremath{76.5} \\
\addlinespace[3pt]
\textbf{Mean $\pm$ SD} & \ensuremath{28.2 \pm 11.4} & \ensuremath{49.0 \pm 15.8} & \ensuremath{74.4 \pm 9.7} \\
\midrule
\multicolumn{4}{@{}l}{\textbf{S2: Kitchen Table}} \\
Blue bowl $\rightarrow$ plate & 75.0 & 84.0 & 86.0 \\
Carrot $\rightarrow$ basket & 90.0 & 94.0 & 98.6 \\
Eggplant $\rightarrow$ basket & 32.3 & 74.0 & 82.5 \\
Red chili $\rightarrow$ basket & 84.0 & 86.0 & 96.2 \\
\addlinespace[3pt]
\textbf{Mean $\pm$ SD} & $70.3 \pm 26.1$ & $84.5 \pm 8.2$ & $90.8 \pm 7.8$ \\
\midrule
\multicolumn{4}{@{}l}{\textbf{S3: Household Table}} \\
Soda bottle $\rightarrow$ shelf & 62.0 & 76.0 & 88.0 \\
Pencil case $\rightarrow$ metal box & 72.0 & 94.0 & 96.0 \\
Snack bag $\rightarrow$ shelf & 28.6 & 68.0 & 75.5 \\
Candy wrapper $\rightarrow$ can & 48.0 & 76.0 & 82.0 \\
\addlinespace[3pt]
\textbf{Mean $\pm$ SD} & $52.7 \pm 18.8$ & $78.5 \pm 11.0$ & $85.4 \pm 8.7$ \\
\midrule
\textbf{Scene-balanced mean} & \ensuremath{50.4} & \ensuremath{70.7} & \ensuremath{83.5} \\
\bottomrule
\end{tabularx}

\end{table}

\subsection{Real-world transfer}
Table~\ref{si-tab:real_tasks} gives the exact physical test counts, including task-level ties and lower observed outcomes. Tests are independent of HIL(human-in-the-loop) training and exclude external correction. Scene-balanced and pooled summaries use different weighting, as defined in Appendix~\ref{si-sec:protocol}.
\begin{table}[!htbp]
\centering
\caption{\textbf{Task-level real-world performance.}
Task entries are successes out of ten autonomous tests.
Scene summaries report mean success (\%) $\pm$ sample SD across subtasks; the final row weights scenes equally, without an SD.
R1/R2 use 200/400 simulated training trajectories per subtask; +10 adds ten real HIL trajectories per subtask.
Real 10/100/200 denote real-only training with 10/100/200 demonstrations per subtask.}
\label{si-tab:real_tasks}
\footnotesize\setlength{\tabcolsep}{3.5pt}\renewcommand{\arraystretch}{1.15}
\begin{tabularx}{\linewidth}{@{}Xccccccc@{}}
\toprule
\textbf{Task} & R1 & R1+10 & R2 & R2+10 & Real 10 & Real 100 & Real 200 \\
\midrule
\multicolumn{8}{@{}l}{\textbf{S1: Office Table}} \\
Glasses $\rightarrow$ stand & 7/10 & 8/10 & 10/10 & 10/10 & 3/10 & 7/10 & 7/10 \\
Red-cap marker $\rightarrow$ green cup & 3/10 & 6/10 & 3/10 & 7/10 & 0/10 & 4/10 & 7/10 \\
Yellow-cap marker $\rightarrow$ basket & 6/10 & 6/10 & 6/10 & 9/10 & 1/10 & 5/10 & 7/10 \\
Mouse $\rightarrow$ mouse pad & 6/10 & 8/10 & 8/10 & 9/10 & 3/10 & 8/10 & 8/10 \\
Red pen $\rightarrow$ green box & 2/10 & 5/10 & 2/10 & 7/10 & 1/10 & 5/10 & 7/10 \\
Black pen $\rightarrow$ pen holder & 2/10 & 5/10 & 3/10 & 7/10 & 0/10 & 5/10 & 6/10 \\
\addlinespace[3pt]
\textbf{Mean $\pm$ SD} & \makecell{43.3\\{\scriptsize $\pm$ 22.5}} & \makecell{63.3\\{\scriptsize $\pm$ 13.7}} & \makecell{53.3\\{\scriptsize $\pm$ 32.0}} & \makecell{81.7\\{\scriptsize $\pm$ 13.3}} & \makecell{13.3\\{\scriptsize $\pm$ 13.7}} & \makecell{56.7\\{\scriptsize $\pm$ 15.1}} & \makecell{70.0\\{\scriptsize $\pm$ 6.3}} \\
\midrule
\multicolumn{8}{@{}l}{\textbf{S2: Kitchen Table}} \\
Blue bowl $\rightarrow$ plate & 5/10 & 9/10 & 6/10 & 10/10 & 5/10 & 6/10 & 9/10 \\
Carrot $\rightarrow$ basket & 6/10 & 8/10 & 7/10 & 7/10 & 5/10 & 7/10 & 7/10 \\
Eggplant $\rightarrow$ basket & 6/10 & 8/10 & 9/10 & 8/10 & 4/10 & 6/10 & 8/10 \\
Red chili $\rightarrow$ basket & 7/10 & 8/10 & 6/10 & 9/10 & 4/10 & 6/10 & 7/10 \\
\addlinespace[3pt]
\textbf{Mean $\pm$ SD} & \makecell{60.0\\{\scriptsize $\pm$ 8.2}} & \makecell{82.5\\{\scriptsize $\pm$ 5.0}} & \makecell{70.0\\{\scriptsize $\pm$ 14.1}} & \makecell{85.0\\{\scriptsize $\pm$ 12.9}} & \makecell{45.0\\{\scriptsize $\pm$ 5.8}} & \makecell{62.5\\{\scriptsize $\pm$ 5.0}} & \makecell{77.5\\{\scriptsize $\pm$ 9.6}} \\
\midrule
\multicolumn{8}{@{}l}{\textbf{S3: Household Table}} \\
Soda bottle $\rightarrow$ shelf & 6/10 & 9/10 & 7/10 & 8/10 & 3/10 & 5/10 & 7/10 \\
Pencil case $\rightarrow$ metal box & 6/10 & 8/10 & 8/10 & 9/10 & 2/10 & 5/10 & 7/10 \\
Snack bag $\rightarrow$ shelf & 5/10 & 8/10 & 6/10 & 9/10 & 2/10 & 6/10 & 9/10 \\
Candy wrapper $\rightarrow$ can & 4/10 & 5/10 & 5/10 & 7/10 & 1/10 & 5/10 & 8/10 \\
\addlinespace[3pt]
\textbf{Mean $\pm$ SD} & \makecell{52.5\\{\scriptsize $\pm$ 9.6}} & \makecell{75.0\\{\scriptsize $\pm$ 17.3}} & \makecell{65.0\\{\scriptsize $\pm$ 12.9}} & \makecell{82.5\\{\scriptsize $\pm$ 9.6}} & \makecell{20.0\\{\scriptsize $\pm$ 8.2}} & \makecell{52.5\\{\scriptsize $\pm$ 5.0}} & \makecell{77.5\\{\scriptsize $\pm$ 9.6}} \\
\midrule
\textbf{Scene-balanced mean} & $51.9$ & $73.6$ & $62.8$ & $83.1$ & $26.1$ & $57.2$ & $75.0$ \\
\bottomrule
\end{tabularx}

\end{table}

\begin{figure}[!htbp]
\centering
\includegraphics[width=.96\linewidth]{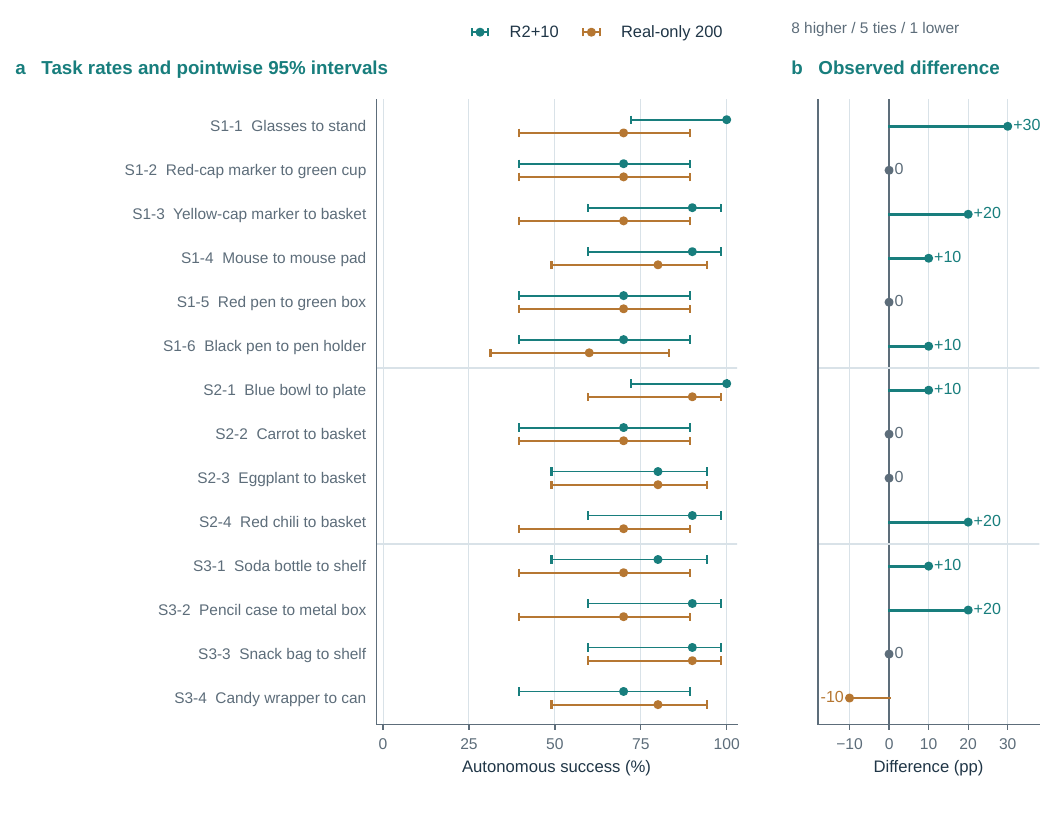}
\caption{\textbf{Task-level real-world transfer and sampling precision.} \textbf{a}, Success rates for R2+10 and Real-only 200, with pointwise 95\% Wilson intervals from ten autonomous tests per task and setting. 
% Intervals condition on each trained policy and assume comparable independent trials; they do not measure training-run variability or simultaneous uncertainty across tasks. 
\textbf{b}, Observed differences matched by subtask identity; positive values favor R2+10. R2+10 uses 400 simulated and ten real-world training trajectories per subtask, versus 200 real-world demonstrations for Real-only 200. Task IDs follow Table~\ref{si-tab:task_definitions}.}
\label{si-fig:task_gains}
\end{figure}

Figure~\ref{si-fig:task_gains} shows the task-level comparison between
R2+10 and Real-only 200, including ties and the negative difference.

\subsection{Matched-trajectory collection comparisons}

\begin{table}[!htbp]
\centering
\caption{\textbf{Scene-level matched-count comparisons.}
\textbf{a}, Simulation evaluation without physical refinement.
\textbf{b}, Real-world performance before and after HIL refinement.
Entries are mean autonomous success (\%) $\pm$ sample SD across subtasks ($n=6$ for S1 and $n=4$ for S3).
$\Delta$ is the mean $\pm$ SD of task-matched differences, in percentage points.
Equal counts do not equate total acquisition cost.}
\label{si-tab:ablation_scene_summary}
\small\setlength{\tabcolsep}{4pt}\renewcommand{\arraystretch}{1.17}
\begin{tabularx}{\linewidth}{@{}llc>{\raggedleft\arraybackslash}X>{\raggedleft\arraybackslash}X>{\raggedleft\arraybackslash}X@{}}
\toprule
\textbf{Scene} & \makecell[l]{\textbf{Sim.}\\\textbf{per subtask}} & \makecell{\textbf{HIL}\\\textbf{per subtask}} & \makecell[r]{\textbf{Static}\\\textbf{collection}} & \makecell[r]{\textbf{Adaptive}\\\textbf{collection}} & \textbf{$\Delta$ (pp)} \\
\midrule
\multicolumn{6}{@{}l}{\textbf{a\quad Simulation evaluation}} \\
S1 & 200 & \NA & \ensuremath{40.7 \pm 14.6} & \ensuremath{49.0 \pm 15.8} & \ensuremath{+8.3 \pm 12.3} \\
 & 400 & \NA & \ensuremath{51.6 \pm 10.2} & \ensuremath{74.4 \pm 9.7} & \ensuremath{+22.9 \pm 9.3} \\
\addlinespace[2pt]
S3 & 200 & \NA & \ensuremath{58.1 \pm 28.1} & \ensuremath{78.5 \pm 11.0} & \ensuremath{+20.5 \pm 18.0} \\
 & 400 & \NA & \ensuremath{67.0 \pm 23.2} & \ensuremath{85.4 \pm 8.7} & \ensuremath{+18.4 \pm 16.0} \\
\midrule
\multicolumn{6}{@{}l}{\textbf{b\quad Real-world evaluation}} \\
S1 & 200 & 0 & $33.3 \pm 21.6$ & $43.3 \pm 22.5$ & $+10.0 \pm 8.9$ \\
 & 200 & 10 & $48.3 \pm 13.3$ & $63.3 \pm 13.7$ & $+15.0 \pm 5.5$ \\
 & 400 & 0 & $45.0 \pm 28.8$ & $53.3 \pm 32.0$ & $+8.3 \pm 11.7$ \\
 & 400 & 10 & $68.3 \pm 16.0$ & $81.7 \pm 13.3$ & $+13.3 \pm 10.3$ \\
\midrule
S3 & 200 & 0 & $40.0 \pm 20.0$ & $52.5 \pm 9.6$ & $+12.5 \pm 12.6$ \\
 & 200 & 10 & $55.0 \pm 17.3$ & $75.0 \pm 17.3$ & $+20.0 \pm 0.0$ \\
 & 400 & 0 & $47.5 \pm 12.6$ & $65.0 \pm 12.9$ & $+17.5 \pm 5.0$ \\
 & 400 & 10 & $67.5 \pm 9.6$ & $82.5 \pm 9.6$ & $+15.0 \pm 5.8$ \\
\bottomrule
\end{tabularx}

\end{table}

Tables~\ref{si-tab:ablation_sim_tasks} and~\ref{si-tab:ablation_real_tasks}
report the complete S1/S3 simulation and physical collection comparisons at 200
and 400 simulated trajectories per subtask. In simulation, the scene-balanced
adaptive advantage is 14.4 percentage points at 200 trajectories and 20.6 points
at 400.
These differences are calculated from the task-level percentages in
Table~\ref{si-tab:ablation_sim_tasks}, with equal weighting of S1 and S3
and rounding only after aggregation.

Adaptive collection is higher on all six S1 subtasks and three of four S3 subtasks
at both budgets. The S3 pencil-case task remains slightly better with Static:
94.8\% versus 94.0\% at 200 trajectories, and 97.2\% versus 96.0\% at 400.
Larger Static datasets raise both scene means, but do not improve every task;
S1 mouse placement decreases from 67.6\% to 59.1\%. The aggregate trends therefore
coexist with task-level exceptions.

\begin{table}[!htbp]
\centering
\caption{\textbf{Task-level simulation collection comparison.} R0 is the shared 50-trajectory-per-subtask initialization. Static and adaptive policies have matched cumulative training budgets per subtask. Scene rows report mean $\pm$ sample SD across subtasks. The final row weights S1 and S3 equally; it is distinct from the three-scene average in the simulation-evolution table.}
\label{si-tab:ablation_sim_tasks}
\small\setlength{\tabcolsep}{5pt}\renewcommand{\arraystretch}{1.15}
\begin{tabularx}{\linewidth}{@{}Xccccc@{}}
\toprule
 & \textbf{Initial} & \multicolumn{2}{c}{\textbf{200 per subtask}} & \multicolumn{2}{c}{\textbf{400 per subtask}} \\
\cmidrule(lr){3-4}\cmidrule(lr){5-6}
\textbf{Task} & R0 (50) & Static & R1 & Static & R2 \\
\midrule
\multicolumn{6}{@{}l}{\textbf{S1: Office Table}} \\
Glasses $\rightarrow$ stand & 44.5 & \ensuremath{32.6} & 64.9 & \ensuremath{41.4} & 80.7 \\
Red-cap marker $\rightarrow$ green cup & \ensuremath{34.0} & \ensuremath{47.2} & \ensuremath{50.1} & \ensuremath{59.7} & \ensuremath{75.3} \\
Yellow-cap marker $\rightarrow$ basket & 26.1 & \ensuremath{30.0} & \ensuremath{30.9} & \ensuremath{38.6} & \ensuremath{57.0} \\
Mouse $\rightarrow$ mouse pad & 33.3 & \ensuremath{67.6} & 70.1 & \ensuremath{59.1} & 85.1 \\
Red pen $\rightarrow$ green box & 17.5 & \ensuremath{35.3} & \ensuremath{36.0} & \ensuremath{48.2} & \ensuremath{72.0} \\
Black pen $\rightarrow$ pen holder & 14.0 & \ensuremath{31.5} & \ensuremath{42.2} & \ensuremath{62.5} & \ensuremath{76.5} \\
\addlinespace[3pt]
\textbf{Mean $\pm$ SD} & \ensuremath{28.2 \pm 11.4} & \ensuremath{40.7 \pm 14.6} & \ensuremath{49.0 \pm 15.8} & \ensuremath{51.6 \pm 10.2} & \ensuremath{74.4 \pm 9.7} \\
\midrule
\multicolumn{6}{@{}l}{\textbf{S3: Household Table}} \\
Soda bottle $\rightarrow$ shelf & 62.0 & \ensuremath{64.0} & 76.0 & \ensuremath{72.0} & 88.0 \\
Pencil case $\rightarrow$ metal box & 72.0 & \ensuremath{94.8} & 94.0 & \ensuremath{97.2} & 96.0 \\
Snack bag $\rightarrow$ shelf & 28.6 & \ensuremath{30.9} & 68.0 & \ensuremath{54.2} & 75.5 \\
Candy wrapper $\rightarrow$ can & 48.0 & \ensuremath{42.5} & 76.0 & \ensuremath{44.4} & 82.0 \\
\addlinespace[3pt]
\textbf{Mean $\pm$ SD} & $52.7 \pm 18.8$ & \ensuremath{58.1 \pm 28.1} & $78.5 \pm 11.0$ & \ensuremath{67.0 \pm 23.2} & $85.4 \pm 8.7$ \\
\midrule
\textbf{Scene-balanced mean} & \ensuremath{40.4} & \ensuremath{49.4} & \ensuremath{63.8} & \ensuremath{59.3} & \ensuremath{79.9} \\
\bottomrule
\end{tabularx}

\end{table}

\begin{table}[!htbp]
\centering
\caption{\textbf{Task-level real-world collection comparison.}
Task entries are successes out of ten autonomous test rollouts.
All budgets are training trajectories per subtask, and one policy covers each scene.
R1 and R2 correspond to simulation budgets of 200 and 400; +10 adds ten real-world HIL training trajectories per subtask for either collection method.
Summary rows report mean success (\%) $\pm$ sample SD across subtasks within each scene.}
\label{si-tab:ablation_real_tasks}
\footnotesize\setlength{\tabcolsep}{3pt}\renewcommand{\arraystretch}{1.15}
\begin{tabularx}{\linewidth}{@{}Xcccccccc@{}}
\toprule
\textbf{Task} & \makecell{Static\\200} & \makecell{Static\\200+10} & R1 & R1+10 & \makecell{Static\\400} & \makecell{Static\\400+10} & R2 & R2+10 \\
\midrule
\multicolumn{9}{@{}l}{\textbf{S1: Office Table}} \\
Glasses $\rightarrow$ stand & 7/10 & 7/10 & 7/10 & 8/10 & 7/10 & 8/10 & 10/10 & 10/10 \\
Red-cap marker $\rightarrow$ green cup & 2/10 & 4/10 & 3/10 & 6/10 & 3/10 & 6/10 & 3/10 & 7/10 \\
Yellow-cap marker $\rightarrow$ basket & 4/10 & 4/10 & 6/10 & 6/10 & 6/10 & 8/10 & 6/10 & 9/10 \\
Mouse $\rightarrow$ mouse pad & 4/10 & 6/10 & 6/10 & 8/10 & 8/10 & 8/10 & 8/10 & 9/10 \\
Red pen $\rightarrow$ green box & 1/10 & 4/10 & 2/10 & 5/10 & 1/10 & 4/10 & 2/10 & 7/10 \\
Black pen $\rightarrow$ pen holder & 2/10 & 4/10 & 2/10 & 5/10 & 2/10 & 7/10 & 3/10 & 7/10 \\
\addlinespace[3pt]
\textbf{Mean $\pm$ SD} & \makecell{33.3\\{\scriptsize $\pm$ 21.6}} & \makecell{48.3\\{\scriptsize $\pm$ 13.3}} & \makecell{43.3\\{\scriptsize $\pm$ 22.5}} & \makecell{63.3\\{\scriptsize $\pm$ 13.7}} & \makecell{45.0\\{\scriptsize $\pm$ 28.8}} & \makecell{68.3\\{\scriptsize $\pm$ 16.0}} & \makecell{53.3\\{\scriptsize $\pm$ 32.0}} & \makecell{81.7\\{\scriptsize $\pm$ 13.3}} \\
\midrule
\multicolumn{9}{@{}l}{\textbf{S3: Household Table}} \\
Soda bottle $\rightarrow$ shelf & 5/10 & 7/10 & 6/10 & 9/10 & 5/10 & 6/10 & 7/10 & 8/10 \\
Pencil case $\rightarrow$ metal box & 5/10 & 6/10 & 6/10 & 8/10 & 6/10 & 8/10 & 8/10 & 9/10 \\
Snack bag $\rightarrow$ shelf & 5/10 & 6/10 & 5/10 & 8/10 & 5/10 & 7/10 & 6/10 & 9/10 \\
Candy wrapper $\rightarrow$ can & 1/10 & 3/10 & 4/10 & 5/10 & 3/10 & 6/10 & 5/10 & 7/10 \\
\addlinespace[3pt]
\textbf{Mean $\pm$ SD} & \makecell{40.0\\{\scriptsize $\pm$ 20.0}} & \makecell{55.0\\{\scriptsize $\pm$ 17.3}} & \makecell{52.5\\{\scriptsize $\pm$ 9.6}} & \makecell{75.0\\{\scriptsize $\pm$ 17.3}} & \makecell{47.5\\{\scriptsize $\pm$ 12.6}} & \makecell{67.5\\{\scriptsize $\pm$ 9.6}} & \makecell{65.0\\{\scriptsize $\pm$ 12.9}} & \makecell{82.5\\{\scriptsize $\pm$ 9.6}} \\
\bottomrule
\end{tabularx}

\end{table}

\subsection{Qualitative execution observations}
\begin{figure}[!t]
\centering
\includegraphics[width=.90\linewidth]{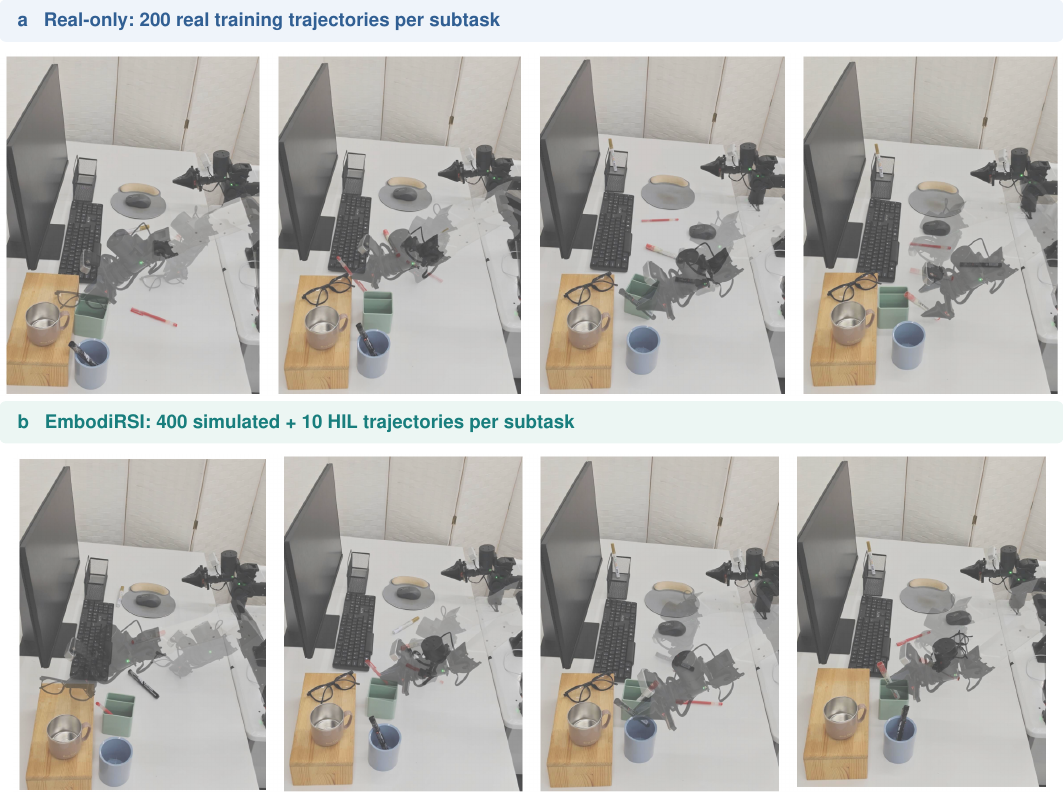}
\caption{\textbf{Illustrative autonomous real-world executions.}
\textbf{a}, Real-only adaptation with 200 real demonstrations per subtask.
\textbf{b}, EmbodiRSI with 400 simulated and ten real HIL trajectories per subtask.
All tests use the learned policy alone.
Quantitative results are reported in Table~\ref{tab:real_main}.}
\label{fig:real_examples}
\end{figure}

The observations below supplement the behavioral observations in Section~\ref{sec:experiments} and the selected sequences in Figure~\ref{fig:real_examples}.
``Reattempt'' denotes an action sequence produced by the learned policy after an unsuccessful interaction, not an agent- or planner-generated training correction.

\paragraph{Simulation errors and instruction-following.}

Missed grasps occurred at peripheral object positions, although these positions were not shown to be unreachable or outside the training distribution.
Placement failures also occurred after successful grasping, particularly when objects missed narrow target openings.
These execution errors are distinct from task-selection errors. At the 50-demonstration initialization, the policy occasionally confused the red-cap-marker and red-pen instructions in S1.
Such confusion was not a recurring failure mode at larger simulation budgets, where most errors arose during execution of the requested manipulation.

Reattempts at R1 and R2 had mixed outcomes: some completed the task, while others failed.
EmbodiRSI's learning workflow combines experience from imperfect intermediate states with additional demonstrations under difficult configurations.

\paragraph{Physical transfer and local corrective experience.}

During HIL collection, operators mainly corrected the final approach, acquisition and placement. After refinement, broad approach motions resembled the simulation demonstrations, while adjustments near grasping and placement resembled the physical corrections. This resemblance does not imply online inverse-kinematics control, nor that HIL changes only contact-related behavior. The pretrained policy's prior capabilities also contribute; not all task knowledge need originate in the simulation data.

Static+10 also benefited from local corrections, but remained less consistent
under pose changes and showed fewer reattempts than the refined evolved policies.

\paragraph{Real-only coverage and task selection.}
With ten real-world demonstrations per subtask, successful behavior was largely
restricted to particular configurations; the non-zero test counts are retained
in Table~\ref{si-tab:real_tasks}. The 200-demonstration collection
covered more configurations than the 100-demonstration collection, although both
policies remained vulnerable at difficult poses and sometimes selected a task
inconsistent with the instruction.

Trained operators followed a standard operating procedure but differed in motion
style and preferred reset configurations. Uneven coverage does not make individual
demonstrations invalid. It may leave difficult configurations sparsely represented,
while correlations among layouts, execution styles and task labels could encourage
reliance on incidental cues instead of the instruction.
\section{Additional related work}
\label{si-sec:related_work}

\subsection{Autonomous improvement and corrective experience}
Autonomous robot learning has explored how experience generated by a
policy can support subsequent policy updates. RoboCat uses
robot-generated data to train successive generalist manipulation
models~\citep{bousmalis2023robocat}, while SOAR employs foundation models to guide autonomous experience collection and evaluation for improving instruction-following skills~\citep{zhou2024autonomous}.
PLD trains residual reinforcement-learning experts, collects recovery trajectories through a hybrid rollout procedure, and distills the resulting data into a VLA model through supervised fine-tuning~\citep{xiao2026self}.
These approaches provide precedents for learning from policy-dependent experience rather, than relying exclusively on an initial demonstration set. Within this line of work, EmbodiRSI focuses on coupling corrective continuations from policy-failed states with updates to subsequent collection configurations.
Its recursion operates through the experience acquired for a fixed policy-learning formulation, rather than through changes to the learning algorithm itself.

\subsection{Failure-driven acquisition and adaptive curricula}
Policy feedback can guide not only corrective supervision but also
the allocation of subsequent training experience. Data-Efficient
Multitask DAgger distributes additional expert demonstrations across
tasks according to policy performance and estimated data
utility~\citep{fu2025data}. Prioritized Level Replay
selects training levels using estimates of their learning
potential~\citep{jiang2021prioritized}. More directly related to failure-driven
generation, Fail2Progress synthesizes simulation data targeted to
observed failures and uses it to refine skill effect
models~\citep{huang2025fail2progress}. RoboPearls uses VLM-based failure analysis to formulate targeted simulation requests for further policy training~\citep{tao2025robopearls}.
EmbodiRSI studies a configuration-level acquisition mechanism: attributed rollout failures guide the selection of source-object and target-anchor reset regions within each task.
The evaluated procedure maintains fixed per-subtask retained-data budgets and mixes targeted configurations with the base collection
distribution.
Its failure-concentration statistics serve as an acquisition heuristic, not an estimate of configuration-specific learning gains.

\subsection{Agentic workflows and execution-time recovery}
Agentic systems operate at different levels of the robotics workflow.
EmbodiedClaw exposes embodied-AI development activities, including environment construction, trajectory synthesis and evaluation, as conversationally executable skills~\citep{zhou2026embodiedclaw}.
ABot-Claw combines persistent cross-embodiment memory with coordinated execution and feedback-based correction~\citep{huo2026abot}, while PhysiAgent integrates monitoring and self-reflection into physical
task execution~\citep{wang2025physiagent}. At the motion-generation
level, ReKep optimizes visually grounded relational keypoint constraints to produce robot actions~\citep{huang2024rekep}. AgentChord augments task graphs with precompiled recovery branches for handling anticipated execution deviations~\citep{xu2026reaction}.
These mechanisms are relevant to workflow coordination and corrective execution, but their roles should be distinguished from the training-data role of recovery in EmbodiRSI.
Our diagnostic and planning tools generate experience during collection; the updated policy is subsequently evaluated without those external recovery mechanisms.
Policy-generated reattempts at test time therefore reflect the learned controller, not an online planner takeover.

\end{document}